\documentclass[10pt,journal,compsoc]{IEEEtran}
\ifCLASSOPTIONcompsoc
  \usepackage[nocompress]{cite}
\else
  \usepackage{cite}
\fi
\ifCLASSINFOpdf
\else
\fi
\usepackage{amsmath}
\usepackage{graphicx}
\usepackage[colorinlistoftodos]{todonotes}
\usepackage[colorlinks=true, allcolors=blue]{hyperref}
\usepackage{amssymb}
\usepackage{float}
\usepackage{bm}
\usepackage{epsfig}
\usepackage{epstopdf}
\usepackage{multirow}
\usepackage{booktabs}
\usepackage{pgffor}
\usepackage{tikz}
\usepackage{enumitem}
\usepackage{algorithm,algorithmic}

\renewcommand{\vec}[1]{\mathbf{#1}}

\begin{document}
%
\title{An Introduction to Image Synthesis with \\ Generative Adversarial Nets}
%
%
%
%


\author{He Huang,
        Philip S. Yu
        and Changhu Wang
\IEEEcompsocitemizethanks{\IEEEcompsocthanksitem He Huang and Philip S. Yu are with the Department of Computer Science, University of Illinois at Chicago, USA. \protect\\
Emails: \{hehuang, psyu\}@uic.edu
\IEEEcompsocthanksitem Changhu Wang is with ByteDance AI Lab, China.\protect\\
Email: wangchanghu@bytedance.com
}} 
\IEEEtitleabstractindextext{%
\begin{abstract}
There has been a drastic growth of research in Generative Adversarial Nets (GANs) in the past few years. Proposed in 2014, GAN has been applied to various applications such as computer vision and natural language processing, and achieves  impressive performance. Among the many applications of GAN, image synthesis is the most well-studied one, and research in this area has already demonstrated the great potential of using GAN in image synthesis. In this paper, we provide a taxonomy of methods used in image synthesis, review different models for text-to-image synthesis and image-to-image translation, and discuss some evaluation metrics as well as possible future research directions in image synthesis with GAN.
\end{abstract}
\begin{IEEEkeywords}
Deep Learning, Generative Adversarial Nets, Image Synthesis, Computer Vision.
\end{IEEEkeywords}}

\maketitle

\IEEEdisplaynontitleabstractindextext

%
\IEEEpeerreviewmaketitle

\IEEEraisesectionheading{\section{Introduction}\label{sec:introduction}}

%
%
%
%
\IEEEPARstart{W}{ith} 
recent advances in deep learning, machine learning algorithms have evolved to such an extent that they can compete and even defeat humans in some tasks, such as image classification on ImageNet \cite{ILSVRC15ImageNet}, playing Go \cite{silver2016alphaGo} and Texas Hold'em poker~\cite{brown2017holdem_poker}. However, we still cannot conclude that those algorithms have true ``intelligence'', since knowing how to do something does not necessarily mean understanding something, and it is critical for a truly intelligent agent to understand its tasks. ``\emph{What I cannot create, I do not understand}'', said the famous physicist Richard Feynman. To put this quote in the case of machine learning, we can say that, for machines to understand their input data,  they need to learn to create the data. The most promising approach is to use generative models that learn to discover the essence of data and find a best distribution to represent it. Also, with a learned generative model, we can even draw samples which are not in the training set but follow the same distribution.



As a new framework of generative model, Generative Adversarial Net (GAN) \cite{goodfellow2014GAN}, proposed in 2014, is able to generate better synthetic images than previous generative models, and since then it has become one of the most popular research areas. A Generative Adversarial Net consists of two  neural networks, a generator and a discriminator, where the generator tries to produce realistic samples that fool the discriminator, while the discriminator tries to distinguish real samples from generated ones. There are two main threads of research on GAN. One is the theoretical thread that tries to alleviate the instability and  mode collapse problems of GAN \cite{radford2015dcgan} \cite{salimans2016improvedGAN} \cite{che2016mrGAN} \cite{arjovsky2017wgan} \cite{gulrajani2017wgan-gp} \cite{berthelot2017began}, or reformulate it from different angles like information theory \cite{chen2016infoGAN} and energy-based models \cite{zhao2016ebGAN}. The other thread focuses on the applications of GAN in computer vision (CV) \cite{radford2015dcgan}, natural language processing (NLP) \cite{yu2017seqgan} and other areas. 

There is a great tutorial given by Goodfellow in NIPS 2016 on GAN~\cite{goodfellow2016gan-tutorial} where he describes the importance of generative models, explains how GAN works, compares GAN with other generative model and discusses frontier research topics in GAN. Also, there is a recent review paper on GAN~\cite{creswell2017gan-survey} which reviews several GAN architectures and training techniques, and introduces some applications of GAN. However, both of these papers are in a general scope, without going into details of specific applications. In this paper, we specifically focus on image synthesis, whose goal is to generate images, since it is by far the most studied area where GAN has been applied. Besides image synthesis, there are many other applications of GAN in computer vision, such as image in-painting~\cite{Yeh2016gan_inpaint}, image captioning~\cite{chen2017showAdpTell}\cite{liang2017rttgan}\cite{zhao2017dualcap}, object detection~\cite{wang2017objdetGAN} and semantic segmentation~\cite{luc2016semanticGAN}.

Research of applying GAN in natural language processing is also a growing trend, such as text modeling~\cite{yu2017seqgan}\cite{Rajeswar2017gan_nl}\cite{Guo2017long_text_gan}, dialogue generation~\cite{li2017gan_dialogue}, question answering~\cite{yang2017semi_gan_qa} and neural machine translation~\cite{yang2017gan_nmt}. However, training GAN in NLP tasks is more difficult and requires more techniques~\cite{yu2017seqgan}, which also makes it a challenging but intriguing research area.

The main goal of this paper is to provide an overview of the methods used in image synthesis with GAN and point out strengths and weaknesses of current methods. We classify the main approaches in image synthesis into three methods, i.e. direct methods, hierarchical methods and iterative methods. Besides these most commonly used methods, there are also other methods which we will briefly mention. We then give a detailed discussion in two of the most important tasks in image synthesis, i.e. text-to-image synthesis and image-to-image translation. We also discuss the possible reasons why GAN performs so well in certain tasks and its role in our goal to artificial intelligence. The goal of this paper is to provide a simple guideline for those who want to apply GAN to their problems and help further research in GAN.


The rest of this paper is organized as follows. In Section \ref{sec:prelim} we first review some core concepts of GAN, as well as some variants and training issues. Then in Section \ref{sec:three-appr} we introduce three main approaches and some other approaches used in image synthesis. In Section \ref{sec:txt2img}, we discuss several methods in text-to-image synthesis and some possible research directions for improvements. In Section \ref{sec:img2img}, we first introduce supervised and unsupervised methods for image-to-image translation, and then turn to more specific applications like face editing, video prediction and image super-resolution. In Section \ref{sec:metrics} we review some evaluation metrics for synthetic images, while in Section \ref{sec:learn_loss} we discuss the discriminator's role as a learned loss function. Conclusions are given in Section~\ref{sec:conclusion}.

\section{GAN Preliminaries}
\label{sec:prelim}
In this section, we review some core concepts of Generative Adversarial Nets (GANs) and some improvements. 

 A generative model $G$ parameterized by $\vec{\theta}$ takes as input a random noise $\vec{z}$ and output a sample $G(\vec{z};\vec{\theta})$, so the output can be regarded as a sample drawn from a distribution: $G(\vec{z};\vec{\theta}) \sim p_g$. Meanwhile, we have a lot of training data $\vec{x}$ drawn from $p_{data}$, and the training objective for the generative model $G$ is to approximate $p_{data}$ using $p_g$.

\begin{figure}[htb]
\centering
\includegraphics[width=0.85\linewidth]{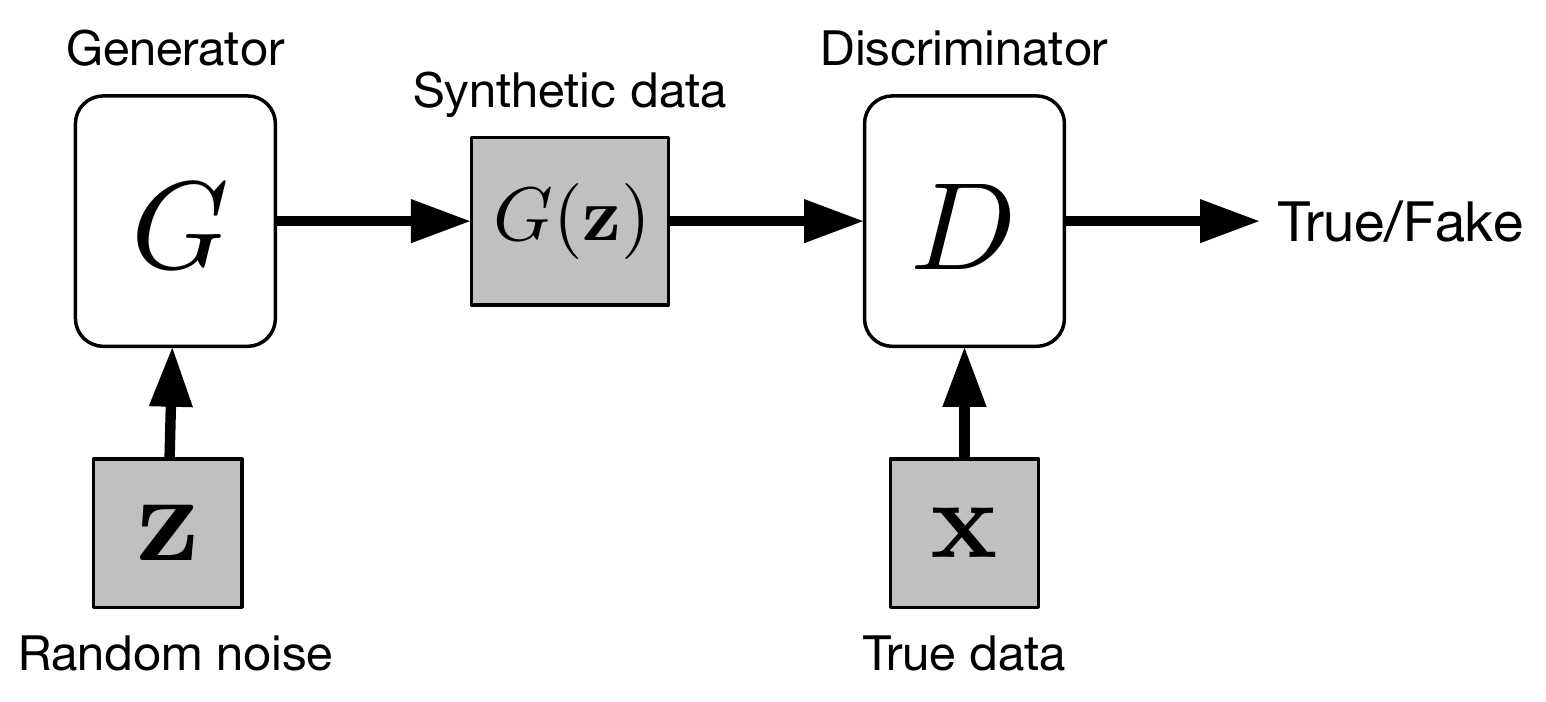}
\caption{General structure of a Generative Adversarial Network, where the generator $G$ takes a noise vector $\vec{z}$ as input and output a synthetic sample $G(\vec{z})$, and the discriminator takes both the synthetic input $G(\vec{z})$ and true sample $\vec{x}$ as inputs and predict whether they are real or fake.}
\label{fig:gan-arch}
\end{figure}

Generative Adversarial Net (GAN)\cite{goodfellow2014GAN} consists of two separate neural networks: a generator $G$ that takes a random noise vector $\vec{z}$, and outputs synthetic data $G(\vec{z})$; a discriminator $D$ that takes an input $\vec{x}$ or $G(\vec{z})$ and output a probability $D(\vec{x})$ or $D(G(\vec{z}))$ to indicate whether it is synthetic or from the true data distribution, as shown in Figure \ref{fig:gan-arch}. Both of the generator and discriminator can be arbitrary neural networks. The first GAN \cite{goodfellow2014GAN} uses fully connected layer as its building block. Later, DCGAN \cite{radford2015dcgan} proposes to use fully convolutional neural networks which achieves better performance, and since then \emph{convolution} and \emph{transposed convolution} layers have become the core components in many GAN models. For more details on (transposed) convolution arithmetic, please refer to this report \cite{dumoulin2016guideCNN}.

The original way to train the generator and discriminator is to form a two-player min-max game where the generator $G$ tries to generate realistic data to fool the discriminator while discriminator $D$ tries to distinguish between real and synthetic data \cite{goodfellow2014GAN}. The value function to be optimized is shown in Equation \ref{eq:gan_obj}, where $p_{data}( \vec{{x}})$ denotes the true data distribution and $p_{z}( \vec{{z}})$ denote the noise distribution. 

\begin{align}
\label{eq:gan_obj}
\min_G \max_D V(D,G) = \ & \mathbb{E}_{ \vec{{x}} \sim p_{data}( \vec{x})}[\log D(\vec{{x}})] \notag \\  & + \mathbb{E}_{ \vec{{z}} \sim p_{z}( \vec{{z}})} [\log(1-D(G(\vec{{z}})))]
\end{align}

However, when the discriminator is trained much better than the generator, $D$ can reject the samples from $G$ with confidence close to 1, and thus the loss $\log(1-D(G(\vec{z})))$ saturates and $G$ can not learn anything from zero gradient. To prevent this, instead of training $G$ to minimize $\log(1-D(G(\vec{z})))$, we can train it to maximize $\log D(G(\vec{z}))$ \cite{goodfellow2014GAN}. Although the new loss function for $G$ gives a different scale of gradient than the original one, it still provides the same direction of gradient and does not saturate.

\subsection{Conditional GAN}
In the original GAN, we have no control of what to be generated, since the output is only dependent on random noise. However, we can add a conditional input $\vec{c}$ to the random noise $\vec{z}$ so that the generated image is defined by $G(\vec{c},\vec{z})$ \cite{mirza2014cGAN}. Typically, the conditional input vector $\vec{c}$ is concatenated with the noise vector $\vec{z}$, and the resulting vector is put into the generator as it is in the original GAN. Besides, we can perform other data augmentation on $\vec{c}$ and $\vec{z}$, as in \cite{zhang2016stackGAN}. The meaning of conditional input $\vec{c}$ is arbitrary, for example, it can be the class of image, attributes of  object \cite{mirza2014cGAN} or an embedding of text descriptions of the image  we want to generate \cite{reed2016txt2img} \cite{reed2016gawwn}.

\subsection{GAN with Auxiliary Classifier}
\begin{figure}[!htb]
\centering
\includegraphics[width=0.9\linewidth]{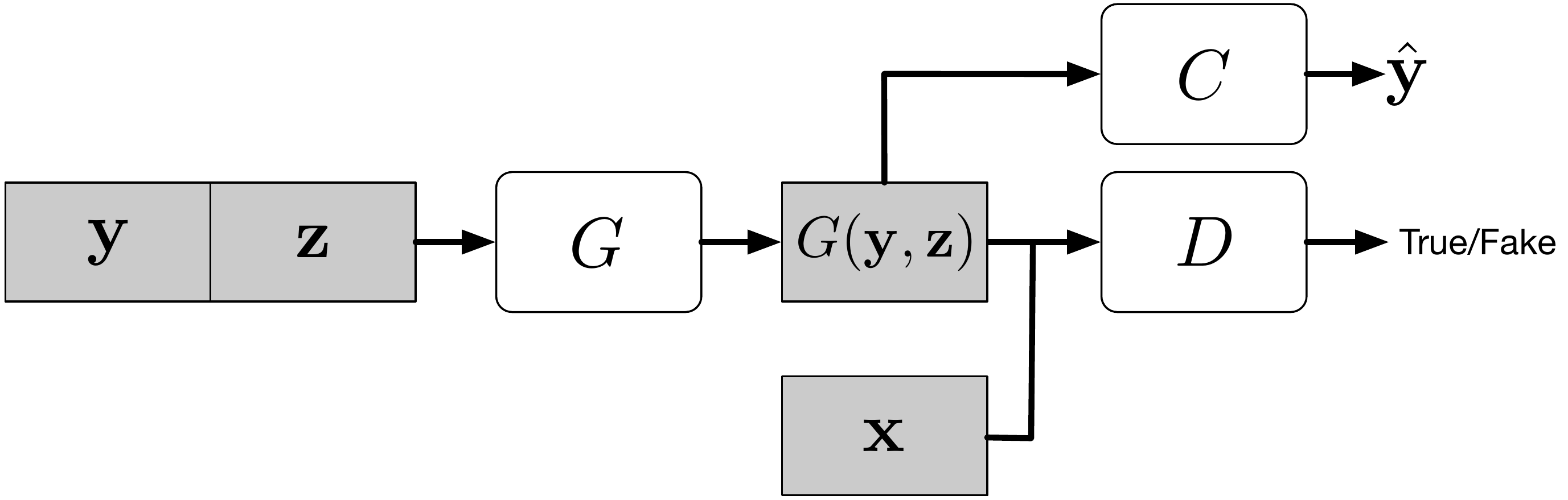}
\caption{Architecture of GAN with auxiliary classifier, where $\vec{y}$ is the conditional input label and $C$ is the classifier that takes the synthetic image $G(\vec{y,z})$ as input and predict its label $\hat{\vec{y}}$}
\label{fig:acgan}
\end{figure}

In order to feed more side-information and to allow for semi-supervised learning, one can add an additional task-specific auxiliary classifier to the discriminator, so that the model is optimized on the original tasks as well as the additional task~\cite{odena2016sgan}~\cite{salimans2016improvedGAN}. The architecture of such method is illustrated in Figure~\ref{fig:acgan}, where $C$ is the auxiliary classifier. Adding auxiliary classifiers allows us to use pre-trained models (e.g. image classifiers trained on ImageNet), and experiments in AC-GAN\cite{odena2016acGAN} demonstrate that such method can help generating sharper images as well as alleviate the \emph{mode collapse} problem. Using auxiliary classifiers can also help in applications such as text-to-image synthesis~\cite{dash2017tacGAN} and image-to-image translation~\cite{yoo2016pldtGAN}.

\subsection{GAN with Encoder}
\begin{figure}[htb]
\centering
\includegraphics[width=0.73\linewidth]{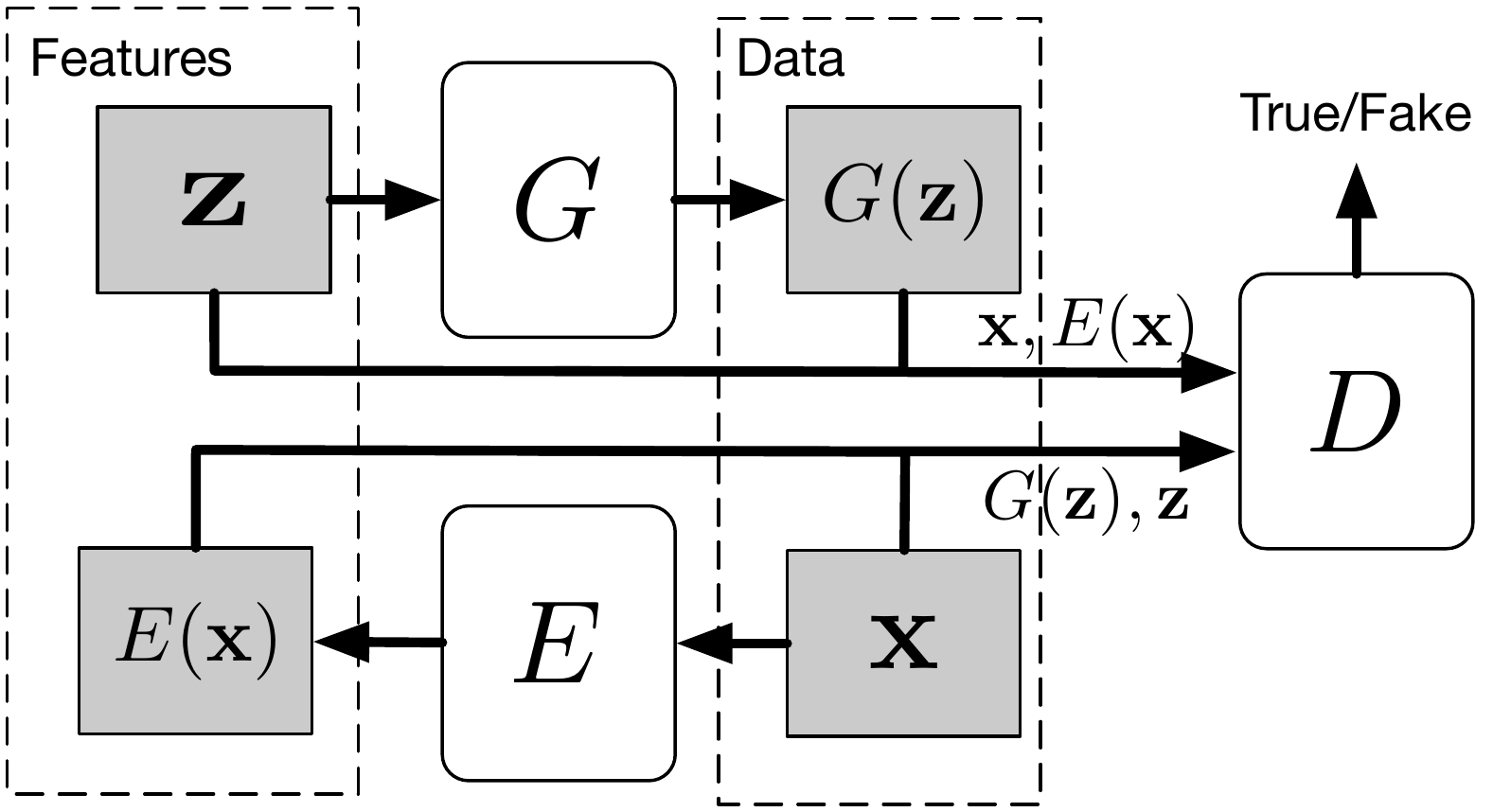}
\caption{Architecture of BiGAN/ALI}
\label{fig:bigan-arch}
\end{figure}

Although GAN can transform a noise vector $\vec{z}$ into a synthetic data sample $G(\vec{z})$, it does not allow inverse transformation. If we treat the noise distribution as a latent feature space for data samples, GAN lacks the ability to map data sample $\vec{x}$ into latent feature $\vec{z}$. In order to allow such mapping, two concurrent works BiGAN~\cite{donahue2016bigan} and ALI~\cite{dumoulin2016ALI} propose to add an encoder $E$ in the original GAN framework, as shown in Figure~\ref{fig:bigan-arch}. Let $\Omega_\vec{x}$ be the data space and $\Omega_\vec{z}$ be the latent feature space, the encoder $E$ takes $\vec{x} \in \Omega_\vec{x}$ as input and produce a feature vector $E(\vec{x}) \in \Omega_\vec{z}$ as output. The discriminator $D$ is modified to take both a data sample and a feature vector as input to calculate $P(Y|\vec{x}, \vec{z})$, where $Y=1$ indicates the sample is real and $Y=0$ means the data is generated by $G$.

The objective is thus defined as:
\begin{align}
    \min_{G,E}\max_{D} V(G,E,D) = &\ \mathbb{E}_{\vec{x} \sim p_{data}(\vec{x})} \log D(\vec{x}, E(\vec{x})) \notag \\
    + &\ \mathbb{E}_{\vec{z} \sim p_{z}(\vec{z})} \log (1-D(G(\vec{z}), \vec{z}))
\end{align}

\subsection{GAN with Variational Auto-Encoder}

\begin{figure}[!htb]
\centering
\includegraphics[width=0.75\linewidth]{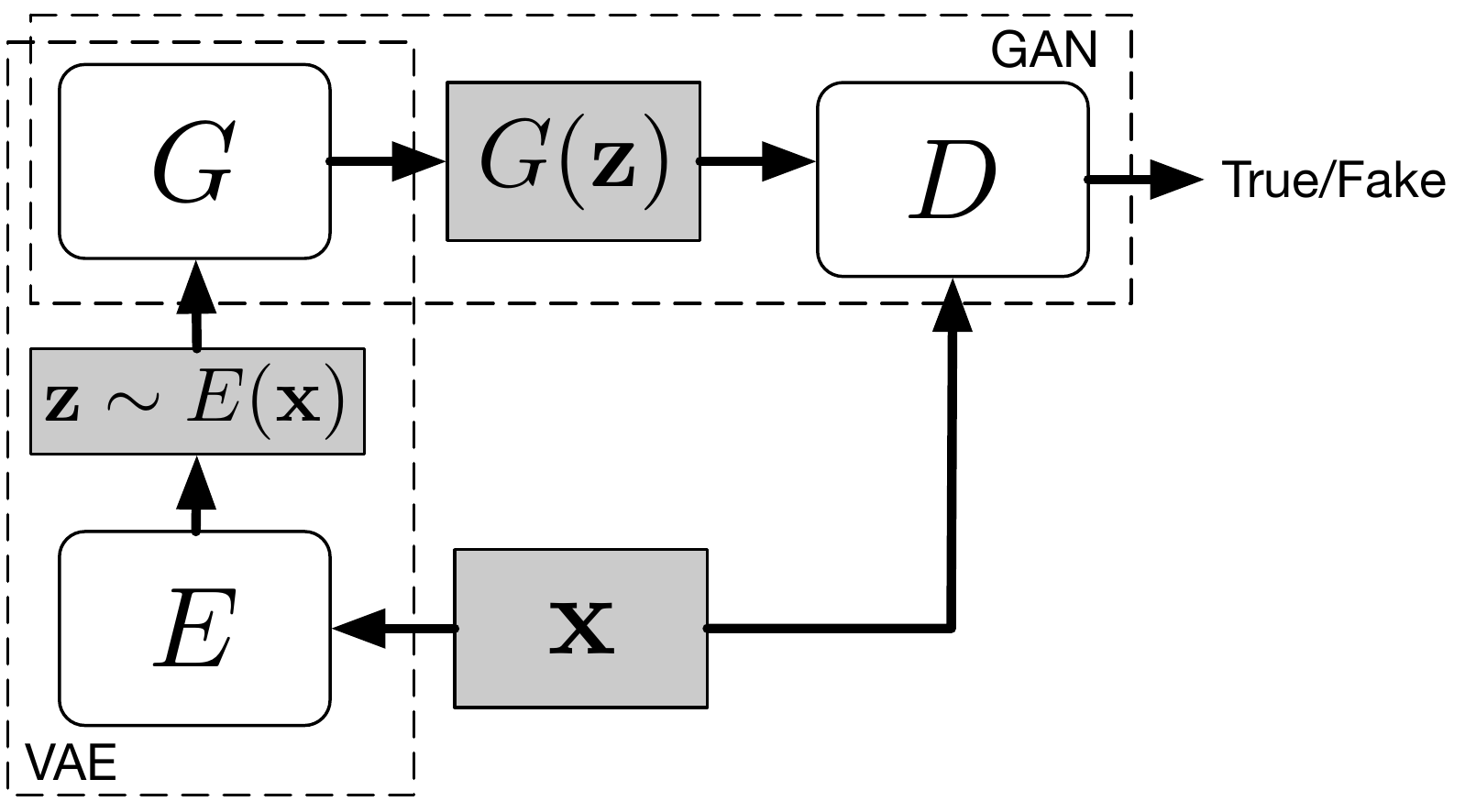}
\caption{Architecture of VAE-GAN}
\label{fig:vae-gan}
\end{figure}

VAE-GAN~\cite{larsen2015vae-gan} proposes to combine Variational Auto-Encoder (VAE)~\cite{kingma2013vae} with GAN~\cite{goodfellow2014GAN} to exploit both of their benefits, as GAN can generate sharp images but often miss some modes while images produced by VAE~\cite{kingma2013vae} are blurry but have large variety. The architecture of VAE-GAN is shown in Figure \ref{fig:vae-gan}. The VAE part regularize the encoder $E$ by imposing a prior of normal distribution (e.g. $\vec{z} \sim \mathcal{N}(0,1)$), and the VAE loss term is defined as:
\begin{align}
    \mathcal{L}_{VAE} = -\mathbb{E}_{\vec{z}\sim q(\vec{z}|\vec{x})} \log[p(\vec{x}|\vec{z})] + D_{\textrm{KL}}(q(\vec{z}|\vec{x})||p(\vec{x})),
\end{align}
where $\vec{z}\sim E(\vec{x}) = q(\vec{z}|\vec{x})$, $\vec{x}\sim G(\vec{z}) = p(\vec{x}|\vec{z})$ and $D_{\textrm{KL}}$ is the Kullback-Leibler divergence.

Also, VAE-GAN~\cite{larsen2015vae-gan} proposes to represent the reconstruction loss of VAE in terms of the discriminator $D$. Let $D_l(\vec{x})$ denotes the representation of the $l$-th layer of the discriminator, and a Gaussian observation model can be defined as:
\begin{align}
    p(D(\vec{x})|\vec{z}) = \mathcal{N}(D(\vec{x})|D(\vec{\tilde{x}}), \mathbf{I}),    
\end{align}
where $\vec{\tilde{x}}\sim G(\vec{z})$ is a sample from the generator, and $\mathbf{I}$ is the identity matrix. So the new VAE loss is:
\begin{align}
\label{eq:vae-gan}
     \mathcal{L}_{VAE} = -\mathbb{E}_{\vec{z}\sim q(\vec{z}|\vec{x})} \log[p(D(\vec{x})|\vec{z})] + D_{\textrm{KL}}(q(\vec{z}|\vec{x})||p(\vec{x})),
\end{align}
which is then combined with the GAN loss defined in Equation \ref{eq:gan_obj}. Experiments demonstrate that VAE-GAN can generate better images than VAE or GAN alone. 

\begin{figure*}[!htb]
\centering
\includegraphics[width=0.85\linewidth]{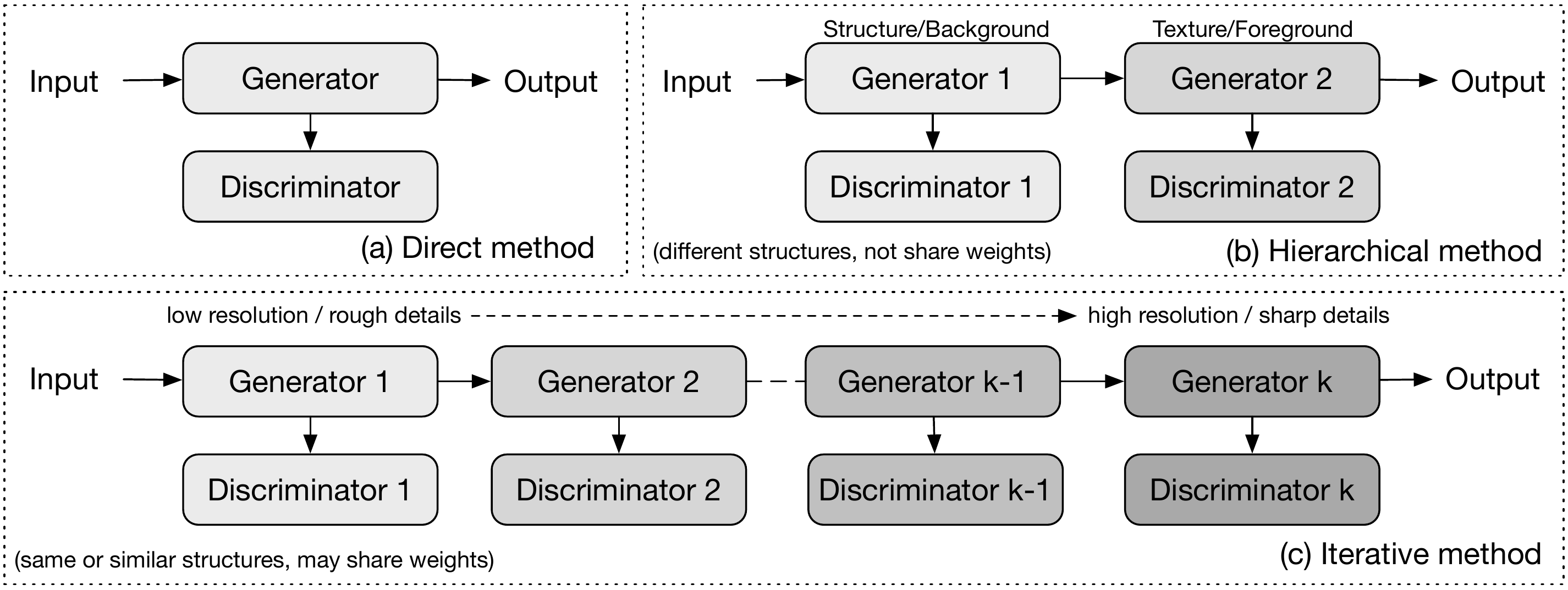}
\caption{Three approaches of image synthesis using Generative Adversarial Networks. The direct method does everything with only one generator and one discriminator, while the other two methods have multiple generators and discriminators. Hierarchical method usually use two layers of GANs, where each GAN plays a fundamentally different role than the other one. Iterative method, however, contains multiple GANs that perform the same task but operate at different resolution.}
\label{fig:gan-methods}
\end{figure*}

\subsection{Handling Mode Collapse}
Although GAN is very effective in image synthesis, its training process is very unstable and requires a lot of tricks to get a good result, as pointed out in \cite{goodfellow2014GAN}\cite{radford2015dcgan}. Despite its instability in training, GAN also suffers from the \emph{mode collapse} problem, as discussed in \cite{goodfellow2014GAN} \cite{radford2015dcgan} \cite{Denton2015lapGAN} . In the original GAN formulation~\cite{goodfellow2014GAN}, the discriminator does not need to consider the variety of synthetic samples, but only focuses on telling whether each sample is realistic or not, which makes it possible for the generator to spend efforts in generating a few samples that are good enough to fool the discriminator. For example, although the MNIST~\cite{lecun1998mnist} dataset contains images of digits from 0 to 9, in an extreme case, a generator only needs to learn to generate one of the ten digits perfectly to completely fool the discriminator, and then the generator stops trying to generate the other nine digits. The absence of the other nine digits is an example of \emph{inter-class mode collapse}. An example of \emph{intra-class mode collapse} is, there are many writing styles for each of the digits, but the generator only learns to generate one perfect sample for each digit to successfully fool the discriminator.

Many methods have been proposed to address the \emph{model collapse} problem. One technique is called \emph{minibatch features}~\cite{salimans2016improvedGAN}, whose idea is to make the discriminator compare an example to a minibatch of true samples as well as a minibatch of generated samples. In this way, the discriminator can learn to tell if a generated sample is too similar to some other generated samples by measuring samples' distances in latent space. Although this method works well, as discussed in \cite{goodfellow2016gan-tutorial}, the performance largely depends on what features are used in distance calculation. MRGAN~\cite{che2016mrGAN} proposes to add an encoder which transforms a sample in data space back to latent space, as  in BiGAN~\cite{donahue2016bigan}. The combination of encoder and generator acts as an auto-encoder, whose reconstruction loss is added to the adversarial loss to act as a mode regularizer. Meanwhile, the discriminator is also trained to discriminate reconstructed samples, which acts as another mode regularizer. WGAN~\cite{arjovsky2017wgan} proposes to use Wasserstein distance to measure the similarity between  true data distribution and the learned distribution, instead of using Jensen-Shannon divergence as in the original GAN~\cite{goodfellow2014GAN}. Although it theoretically avoids mode collapse, it takes a longer time for the model to converge than previous GANs. To alleviate this problem, WGAN-GP~\cite{gulrajani2017wgan-gp} proposes to use gradient penalty, instead of weight clipping in WGAN. WGAN-GP generally produces good images and greatly avoid mode collapse, and it is easy to apply this training framework to other GAN models. More tips for training GANs can be found in Soumith's NIPS 2016 tutorial ``How to train a GAN''\footnote{\url{https://github.com/soumith/ganhacks}}.

\begin{figure}[!htb]
\centering
\includegraphics[width=0.6\linewidth]{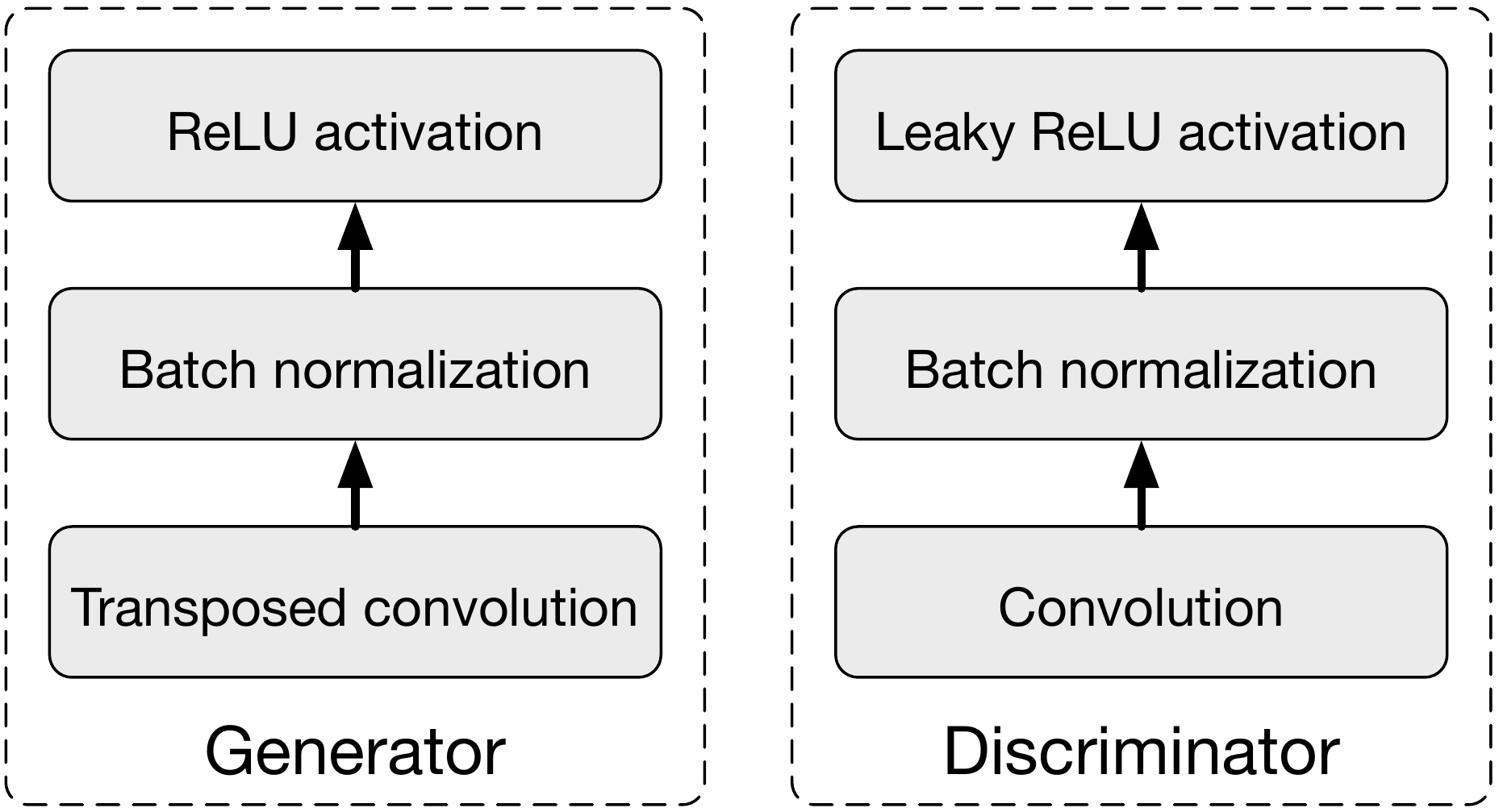}
\caption{Building blocks of DCGAN, where the generator uses transposed convolution, batch-normalization and ReLU activation, while the discriminator uses convolution, batch-normalization and LeakyReLU activation}
\label{fig:dcgan}
\end{figure}

\section{General Approaches of Image Synthesis with GAN}
\label{sec:three-appr}

In this section, we summarize the three main approaches used in generating images, i.e. \emph{direct methods}, \emph{iterative methods} and \emph{hierarchical methods} respectively, which form the basis of all applications mentioned in this paper. The overall structures of these three methods are shown in Figure \ref{fig:gan-methods}.

\subsection{Direct Methods}
All methods under this category follows the philosophy of using one generator and one discriminator in their models, and the structures of the generator and the discriminator are straight-forward without branches. Many of the earliest GAN models fall into this category, like GAN \cite{goodfellow2014GAN}, DCGAN\cite{radford2015dcgan}, ImprovedGAN \cite{salimans2016improvedGAN}, InfoGAN \cite{chen2016infoGAN}, f-GAN~\cite{nowozin2016fGAN} and GAN-INT-CLS \cite{reed2016txt2img}. Among them, DCGAN is one of the most classic ones whose structure is used by many later models such as  \cite{chen2016infoGAN} \cite{reed2016txt2img} \cite{perarnau2016icGAN}  \cite{zhao2017dualcap}. The general building blocks used in DCGAN are shown in Figure \ref{fig:dcgan}, where the generator uses transposed convolution, batch-normalization and ReLU activation, while the discriminator uses convolution, batch-normalization and LeakyReLU activation. 

This kind of method is relatively more straight-forward to design and implement when compared with \emph{hierarchical} and \emph{iterative} methods, and it usually achieves good results.

\subsection{Hierarchical Methods}
Contrary to the \emph{Direct Method}, algorithms under the \emph{Hierarchical Method} use two generators and two discriminators in their models, where different generators have different purposes. The idea behind those methods is to separate an image into two parts, like  ``styles \& structure'' and ``foreground \& background''. The relation between the two generators may be either parallel or sequential. 

SS-GAN~\cite{wang2016ssGAN} proposes to use two GANs, a Structure-GAN for generating a surface normal map from random noise $\hat{\vec{z}}$, and another Style-GAN that takes both the generated surface normal map as well as a noise $\tilde{\vec{z}}$ as input and outputs an image. The Structure-GAN uses the same building blocks as DCGAN~\cite{radford2015dcgan}, while the Style-GAN is slightly different. For Style-Generator, the generated surface normal map and the noise vector go through several convolutional and transposed convolutional layers respectively, and then the results are concatenated into a single tensor which will go through the remaining layers in Style-Generator. As for the Style-Discriminator, each surface normal map and its corresponding image are concatenated at the channel dimension to form a single input to the discriminator. Besides, SS-GAN assumes that, a good synthetic image should also be used to reconstruct a good surface normal map. Under this assumption, SS-GAN designs a fully-connected network that transforms an image back to its surface normal map, and uses a pixel-wise loss that enforces the reconstructed surface normal to approximate the true one. A main limitation of SS-GAN is that it requires to use Kinect to obtain groundtruth for surface normal maps.

As a special example, LR-GAN \cite{yang2017lrGAN} chooses to generate the foreground and background content using different generator, but only one discriminator is used to judge the images while the recurrent image generation process is related to the iterative method. Nonetheless, experiments of LR-GAN demonstrate that it is possible to separate the generation of foreground and background content and produce sharper images.

\subsection{Iterative Methods}
This method differentiates itself from \emph{Hierarchical Methods} in two ways. First, instead of using two different generators that perform different roles, the models in this category use multiple generators that have similar or even the same structures, and they generate images from coarse to fine, with each generator refining the details of the results from the previous generator. Second, when using the same structures in the generators, \emph{Iterative Methods} can use weight-sharing among the generators \cite{yang2017lrGAN}, while \emph{Hierarchical Methods} usually can not. 

LAPGAN \cite{Denton2015lapGAN} is the first GAN that uses an iterative method to generate images from coarse to fine using Laplacian pyramid~\cite{burt1987laplacianPy}. The multiple generators in LAPGAN perform the same task: takes an image from previous generator and a noise vector as input, and then outputs the details (a residual image) that can make the image sharper when added to the input image. The only difference in the structures of those generators is the size of input/output dimension, while an exception is that the generator at the lowest level only takes a noise vector as input and outputs an image. LAPGAN outperforms the original GAN \cite{goodfellow2014GAN} and shows that iterative method can generate sharper images than direct method.

StackGAN \cite{zhang2016stackGAN}, as an iterative method, has only two layers of generators. First generator takes an input $(\vec{z},\vec{c})$ and then outputs a blurry image that can show a rough shape and blurry details of the objects, while the second generator takes $(\vec{z},\vec{c})$ and the image generated by the previous generator and then output a larger image with more photo-realistic details. 

Another example of \emph{Iterative Methods} is SGAN \cite{huang2016sGAN} which stacks generators that takes lower level feature as input and outputs higher level features, while the bottom generator takes a noise vector as input and the top generator outputs an image. The necessity of using separate generators for different levels of features is that SGAN associates an encoder, a discriminator and a Q-network \cite{huang2016sGAN} (which is used to predict the posterior probability $P(\vec{z}_i|\vec{h}_i)$ for entropy maximization, where $\vec{h}_i$ is the output feature of the $i$-th layer) for each generator, so as to constrain and improve the quality of those features. 

An example of using weight-sharing is the GRAN \cite{ImKJM2016gran} model, which is an extension to the  DRAW \cite{gregor2015draw} model which is based on variational autoencoder~\cite{kingma2013vae}. As in DRAW, GRAN generates an image in a recurrent way that feeds the output of the previous step into the model and the output of the current step will be fed back as the input in the next step. All steps use the same generator, so the weights are shared among them, just like classic Recurrent Neural Network (RNN).

\subsection{Other Methods}
PPGN \cite{nguyen2016ppgn} produces impressive images in several tasks, such as class-conditioned image synthesis~\cite{mirza2014cGAN}, text-to-image synthesis~\cite{reed2016txt2img} and image inpainting~\cite{Yeh2016gan_inpaint}. Different from other methods mentioned earlier, PPGN uses \emph{activation maximization} \cite{nguyen2016dgn-am} to generate images, and it is based on sampling with a prior learned with denoising autoencoder (DAE) \cite{vincent2008DAE}. To generate an image conditioned on a certain class label $y$, instead of using a feed-forward way (e.g. recurrent methods can be seen as feed-forward if unfolded through time), PPGN runs an optimization process that finds an input $\vec{z}$ to the generator that makes the output image highly activate a certain neuron in another pretrained classifier (in this case, the neuron in the output layer that corresponds to its class label $y$).

In order to generate better higher resolution images, ProgressiveGAN~\cite{karras2017progressive} proposes to start with training a generator and discriminator of $4\times4$ pixels, after which it incrementally adds extra layers that doubles the output resolution up to $1024\times1024$. This approach allows the model to learn coarse structure first and then focus on refining details later, instead of having to deal with all details at different scale simultaneously.




\section{Text-to-Image Synthesis}
\label{sec:txt2img}
When we apply GAN to image synthesis, it is desired to control the content of generated images. Although there are label-conditioned GAN models like cGAN \cite{mirza2014cGAN} which can generate images belong to a specific class, it remains a great challenge to generate images based on text descriptions. \emph{Text-to-image synthesis} is kind of the holy grail of computer vision, since if an algorithm can generate truly realistic images from mere text descriptions, we can have a high confidence that the algorithm actually understands what is in the images, where computer vision is about teaching computers to see and understand visual contents in the real world.

GAN-INT-CLS~\cite{reed2016txt2img} provides the first attempt of using GAN to generate images from text descriptions. The idea is similar to conditional GAN that concatenates the condition vector with the noise vector, but with the difference of using the embedding of text sentences instead of class labels or attributes. The embedding method used in GAN-INT-CLS~\cite{reed2016txt2img} is from another paper \cite{reed2016learnDescription} that tries to learn robust embeddings of images and sentences given the image-sentence pairs. Except for the sentence embedding method, the generator of GAN-INT-CLS follows the same architecture of DCGAN \cite{radford2015dcgan}. As for the discriminator, in order to take into account the text description, the text embedding vector of length $K$ is \emph{spatially replicated} to become a text embedding tensor of shape $[W \times H \times K]$, where $W$ and $H$ are the width and height of the generated image's feature tensor after going through several convolutional layers in the discriminator. Then the text embedding tensor is combined with the image feature tensor of shape $[W \times H \times C]$ at the channel dimension $C$ and forms a new tensor of shape $[W \times H \times (C+K)]$, which then goes through the rest layers of the discriminator. The intuition behind this approach is that, by spatial replication and depth concatenation, each ``\emph{pixel}'' (of shape $[1\times1\times(C+K)]$) of the image's feature tensor contains all the information of text description, and then the convolutional layers can learn to align the image's content with certain parts of the text feature by using multiple convolution kernels.

GAN-INT-CLS~\cite{reed2016txt2img} also proposes to distinguish between two sources of errors: \emph{unrealistic image with any text}, and \emph{realistic image with mismatched text}. To train the discriminator to distinguish these two kinds of errors, three types of input are fed into the discriminator at every training step: \{\emph{real image, right text}\}, \{\emph{real image, wrong text}\} and \{\emph{fake image, right text}\}. The experimental results in \cite{reed2016txt2img} show that such training technique is important in generating high quality images, since it tells the model not only how to generate realistic images, but also the correspondence between text and images.

In addition, GAN-INT-CLS~\cite{reed2016txt2img} proposes to use manifold interpolation to obtain more text embeddings, by simply interpolating between captions in the training set. Since the number of captions for each image is usually no more than five and an image can be described in many ways, doing such interpolation allows the model to learn from possible text descriptions that are not in the training set. According to the authors, the interpolated text embeddings need not correspond to actual human-written text, so there is no extra labeling required.

TAC-GAN~\cite{dash2017tacGAN} is a combination of GAN-INT-CLS~\cite{reed2016txt2img} and AC-GAN~\cite{odena2016acGAN}. With the auxiliary classifier, TAC-GAN is able to achieve higher \emph{Inception Score}~\cite{salimans2016improvedGAN} than GAN-INT-CLS~\cite{reed2016txt2img} and StackGAN~\cite{zhang2016stackGAN} on the Oxford-102~\cite{Nilsback08oxfordFlower} daataset.


\subsection{Text-to-Image with Location Constraints}
Although GAN-INT-CLS~\cite{reed2016txt2img} and StackGAN~\cite{zhang2016stackGAN} can generate images based on text description, they fail to capture the localization constraints of the objects in images. To allow encoding spatial constraints, Reed et al. propose GAWWN~\cite{reed2016gawwn} that presents two possible solutions.  

The first approach proposed in GAWWN~\cite{reed2016gawwn} is to learn a bounding box for an object by putting a spatially replicated text embedding tensor through a \emph{Spatial Transformer Network}~\cite{jaderberg2015stn}. Here the \emph{spatial replication} is the same process mentioned in GAN-INT-CLS~\cite{reed2016txt2img}. The output of spatial transformer network is a tensor with the same dimension as input, but values outside of the bounding are all zeros.  The output tensor of the spatial transformer goes through several convolutional layers to reduce its size back to a 1-dimensional vector, which not only preserves the information of text but also provides a constraint on object's location by the bounding box. A benefit of this approach is that it is end-to-end, without requiring additional input.

The second approach proposed in GAWWN \cite{reed2016gawwn} is to use user-specified keypoints to constrain the different parts (e.g. head, leg, arm, tail, etc.) of the object in the image. For each keypoint, a mask matrix is generated where the keypoint position is 1 and others 0, and all the matrices are combined through depth concatenation to form a mask tensor of shape $[M \times M \times K]$, where $M$ is the size of the mask and $K$ is the number of keypoints. The tensor is then flattened into a binary matrix with 1 indicating the presence of a keypoint and 0 otherwise, and then replicated depth-wise to become a tensor to be fed into remaining layers. Although this approach allows more detailed constraints on the object, it requires extra user input to specify the keypoints. Even if GAWWN can infer unobserved keypoints from a few user-specified keypoints so that the user does not need to specify all keypoints, the cost of extra user input is still non-trivial.

The remaining part of GAWWN \cite{reed2016gawwn} is similar to GAN-INT-CLS~\cite{reed2016txt2img}, with the difference of using a separate path-way to process bounding box tensor or keypoint tensor. Although GAWWN provides two approaches that can enforce location constraints on the generated images, it only works on images with single objects, since neither of the proposed methods is able to handle several different objects in an image. From the result shown in \cite{reed2016gawwn}, GAWWN works well on the CUB dataset~\cite{wah2011cub_bird}, while the synthetic images generated from the model trained on MPII Human Pose (MHP) dataset~\cite{andriluka2014MHP2} are very blurry and it is hard to tell what the content is. This may be due to the fact that the poses of a standing bird are very similar to each other (note that birds in the CUB dataset are at standing poses), while a human's poses can be of uncountable types. 

The main benefit of specifying the locations of each part of the objects is that it yields more interpretable results, and that it is desirable that the model can understand the concepts of different parts of objects, which is one of the ultimate goals of computer vision.

\subsection{Text-to-Image with Stacked GANs}
Instead of using only one generator, StackGAN~\cite{zhang2016stackGAN} proposes to use two different generators for text-to-image synthesis. The first generator is responsible for generating low-resolution images that contain rough shapes and colors of objects, while the second generator takes the output of the first generator and produces images with higher resolution and sharper details. Each generator is associated with its own discriminator. Besides, StackGAN also proposes a conditional data augmentation technique to produce more text embeddings for the generator. StackGAN randomly samples from a Gaussian distribution $\mathcal{N}(\mu(\vec{\phi}_t),\Sigma(\vec{\phi}_t))$, where the mean vector $\mu(\vec{\phi}_t)$ and diagonal variance matrix $\Sigma(\vec{\phi}_t)$ are functions of text embedding $\vec{\phi}_t$. To further enforce the smoothness over conditional manifold, $\mathcal{N}(\mu(\vec{\phi}_t),\Sigma(\vec{\phi}_t))$ is constrained to approximate a standard Gaussian distribution $\mathcal{N}(0,1)$ by adding a  Kullback-Leibler divergence regularization term. 

As an improved version of StackGAN, StackGAN++~\cite{zhang2017stackGANpp} proposes to have more pairs of generators and discriminators instead of just two, adds an unconditional image synthesis loss to the discriminators, and uses a color-consistency regularization term calculated by mean-square loss of the means and variances between real and fake images.

AttnGAN~\cite{xu2017attnGAN} further extends the architecture of StackGAN++~\cite{zhang2017stackGANpp} by using attention mechanism over image and text features. In AttnGAM, each sentence is embedded into a global sentence vector and each word of the sentence is also embedded into a word vector. The global sentence vector is used to generate a low resolution image at the first stage, and then the following stages use the input image features from the previous stage and the word vectors as input to the attention layer and calculate a word-context vector which will be combined with the image features and form the input to the generator that will generate new image features. Besides, AttnGAN also proposes a Deep Attentional Multimodal Similarity Model (DAMSM) that uses attention layers to compute the similarity between images and text using both global sentence vectors as well as fine-grained word vectors, which provides an additional fine-grained image-text matching loss for training the generator.  Experiments of AttnGAN not only show the effectiveness of using attention layers in image synthesis, but also make the model more interpretable.

With stacked generators, StackGAN~\cite{zhang2016stackGAN}, StackGAN++~\cite{zhang2017stackGANpp} and AttnGAN~\cite{xu2017attnGAN} produce sharper images than GAN-INT-CLS~\cite{reed2016txt2img} and GAWWN~\cite{reed2016gawwn} on the  CUB~\cite{wah2011cub_bird} and Oxford-102~\cite{Nilsback08oxfordFlower} datasets. Although AttnGAN~\cite{xu2017attnGAN} claims to have significantly higher \emph{Inception Score}~\cite{salimans2016improvedGAN} than PPGN~\cite{nguyen2016ppgn} on the COCO~\cite{lin2014mscoco} dataset, the examples it provides do not visually look apparently better.

\subsection{Text-to-Image by Iterative Sampling}
Different from previous approaches that directly incorporate the text information in the generation process, PPGN~\cite{nguyen2016ppgn} proposes to use \emph{activation maximization}~\cite{nguyen2016dgn-am} method to generate images in an iterative sampling way. The model contains two separate parts: a pretrained image captioning model, and an image generator. The image generator is a combination of denoising auto-encoder and GAN, trained independent of the image captioning model. Let $p(\vec{x})$ be the distribution of images, and $p(\vec{y})$ be the distribution of text descriptions, we want to sample image from the joint distribution $p(\vec{x},\vec{y})$. PPGN factorizes the joint distribution into two factors: $p(\vec{x},\vec{y})=p(\vec{x})p(\vec{y}|\vec{x})$, where $p(\vec{x})$ is modeled by the generator, and $p(\vec{y}|\vec{x})$ is modeled by the image captioning model. According to the paper \cite{nguyen2016ppgn}, given a trained image generator and image captioning model, the following iterative sampling process is performed to obtain an image $\vec{x}$ based on the description $\vec{y}^{*}$:

\begin{align}
    \vec{x}_{t+1} = \vec{x}_t + \epsilon_1\frac{\partial \log p(\vec{x}_t)}{\partial \vec{x}_t} + \epsilon_2\frac{\partial \log p(\vec{y}=\vec{y}^{*}|\vec{x}_t)}{\partial \vec{x}_t} + N(0, \epsilon_3^2),
\end{align}

where the $\epsilon_1$ term takes a step from current image $\vec{x}_t$ to a more realistic image (regardless of the text description), the $\epsilon_2$ term takes a step from current image to an image that better matches the description $\vec{y}^{*}$ (here the LRCN model~\cite{donahue2015lrcn} is used for the $p(\vec{y}=\vec{y}^{*}|\vec{x})$ term), and the $\epsilon_3$ term adds a small noise to allow for a broader search in the latent space.

Although such iterative sampling method takes more time to generate an image in test phrase, it can generate higher resolution images with better quality than previous methods like \cite{reed2016txt2img}~\cite{reed2016gawwn}~\cite{zhang2016stackGAN}, and its performance is among the best in both class-conditioned and text-conditioned image synthesis.

\subsection{Limitations of Current Text-to-Image Models}
Present text-to-image models perform well on datasets with single object per image, such as human faces in CelebA \cite{liu2015celebA}, birds in CUB \cite{wah2011cub_bird}, flowers in Oxford-102 \cite{Nilsback08oxfordFlower}, and some objects in ImageNet \cite{ILSVRC15ImageNet}. Also, they can synthesize reasonable images for scenes like bedrooms and living rooms in the LSUN \cite{yu15lsun}, even though the objects in the scenes lack sharp details. However, all present models work badly in situation where multiple complicated objects are involved in one image, as it is in the MS-COCO dataset \cite{lin2014mscoco}.   

A plausible reason why current models fail to work well on complicated images is that the models only learn the overall features of an image, instead of learning the concept of each kind of objects in it. This gives an explanation why synthetic scenes of bedrooms and living rooms lack sharp details, since the model do not distinguish between a bed and a desk, all it sees is that some patterns of shapes and colors should be put somewhere in the synthetic image. In other words, the model does not really understand the image, but just remembers where to put some shapes and colors.

Generative Adversarial Network certainly provides us a promising way to do text-to-image synthesis, since it produces sharper images than any other generative methods so far. To take a further step in text-to-image synthesis, we need to figure out novel ways to teach the algorithms the concepts of things. One possible way is to train separate models that can generate different kinds of objects, and then train another model that learns how to combine different objects (i.e. the reasonable relations between objects) into one image based on the text descriptions. However, such method requires large training sets for different objects, and another large dataset that contains images of those different objects, which is hard to acquire. Another possible direction may be to make use of the \emph{Capsule} idea proposed by Hinton et al. \cite{hinton2011transforming}\cite{sabour2017capsnet} since \emph{capsules} are designed to capture the concepts of objects, but how to efficiently train such \emph{capsule}-based network is still a problem to be solved.

\section{Image-to-Image Translation}
\label{sec:img2img}
In this section, we discuss another type of image synthesis, \emph{image-to-image translation}, which takes images as conditional input.
\emph{Image-to-image translation} is defined as the problem of translating a possible representation of one scene into another, such as mapping BW images into RGB images, or the other way around \cite{isola2016pix2pix}. This problem is related to \emph{style transfer}~\cite{johnson2016perceptual}~\cite{Jing2017nst-survey}, which takes a content image and a style image and output an image with the content of the content image and the style of the style image. \emph{Image-to-image translation} can be viewed as a generalization of \emph{style transfer}, since it is not limited to transferring the styles of images, but can also manipulate attributes of objects (as in applications of \emph{face editing}). In this section, we introduce several models that work for general \emph{image-to-image} translation, from supervised methods to unsupervised ones. The ``supervision'' here means that for each image in the source domain, there is a corresponding ground-truth image in the target domain. Later we will turn into different specific applications such as \emph{face editing}, \emph{image super resolution}, \emph{image in-painting} and \emph{video prediction}. 

\subsection{Supervised Image-to-Image Translation}
Pix2Pix~\cite{isola2016pix2pix} proposes to combine the loss of a conditional Generative Adversarial Network (cGAN) with L1 regularization loss, so that the generator is not only trained to fool the discriminator but also to generate images as close to ground-truth as possible. The reason for using L1 instead of L2 is that L1 produces less blurry images~\cite{Zhao2015lossIP}.

The conditional GAN loss is defined as:
\begin{align}
    \mathcal{L}_{cGAN}(G,D) = &\ \mathbb{E}_{\vec{x},\vec{y}\sim p_{data}(\vec{x},\vec{y})}[\log D(\vec{x},\vec{y})] + \notag \\ 
                         &\ \mathbb{E}_{\vec{x} \sim  p_{data}(\vec{x}), \vec{z} \sim p_z(\vec{z})}[\log(1-D(\vec{x}, G(\vec{x},\vec{z}))], 
\end{align}
where $x,y \sim p(x,y)$ are images of the same scene with different styles, $z\sim p(z)$ is a random noise as in the regular GAN\cite{goodfellow2014GAN}.

The L1 loss for constraining self-similarity is defined as:
\begin{align}
    \label{eq:pix2pix-L1}
     \mathcal{L}_{L1}(G) = \mathbb{E}_{\vec{x},\vec{y}\sim p_{data}(\vec{x},\vec{y}), \vec{z} \sim p_z(\vec{z})}[||\vec{y}-G(\vec{x},\vec{z})||_1],
\end{align}

The overall objective is thus given by:
\begin{align}
    G^*,D^* = \textrm{arg} \min_G \max_D \mathcal{L}_{cGAN}(G,D) + \lambda \mathcal{L}_{L1}(G),
\end{align}
where $\lambda$ is a hyper-parameter to balance the two loss terms.

The authors of Pix2Pix \cite{isola2016pix2pix} find that the noise $z$ does not have obvious effect on the output, so they provide the noise in the form of dropout at training and test time instead of drawing  samples from a random distribution.

The generator structure for Pix2Pix~\cite{isola2016pix2pix} is based on U-Net~\cite{ronneberger2015u-net}, which belongs to the encoder-decoder framework but adds skip connections from encoder to decoder so as to allow circumventing the bottleneck for sharing low-level information like edges of objects. 



Pix2Pix~\cite{isola2016pix2pix} proposes PatchGAN (the patch-based idea was previously explored in MGAN~\cite{li2016mGAN}) as the discriminator, which, instead of classifying the whole image, tries to classify each $N \times N$ path of the image and average all the scores of patches to get the final score for the image. The motivation of this method is that, although L1 and L2 losses produce blurry images and fail to capture high frequency details, in many cases they can capture the low frequencies quite well. In order to capture the high frequency details, Pix2Pix~\cite{isola2016pix2pix} argues that it is sufficient to restrict the discriminator to focus only on local patches. From the experiments, it is found that, for an $256 \times 256$ image, a patch-size of $70 \times 70$ works best. 

Although Pix2Pix produces very impressive synthetic images, the major limitation is that it must use paired images as supervision, as is shown in Equation \ref{eq:pix2pix-L1} that data pair $(\vec{x},\vec{y})$ is drawn from the joint distribution $p(\vec{x},\vec{y})$.

\subsection{Supervised Image-to-Image Translation with Pair-wise Discrimination}
\begin{figure}[htb]
\centering
\includegraphics[width=0.95\linewidth]{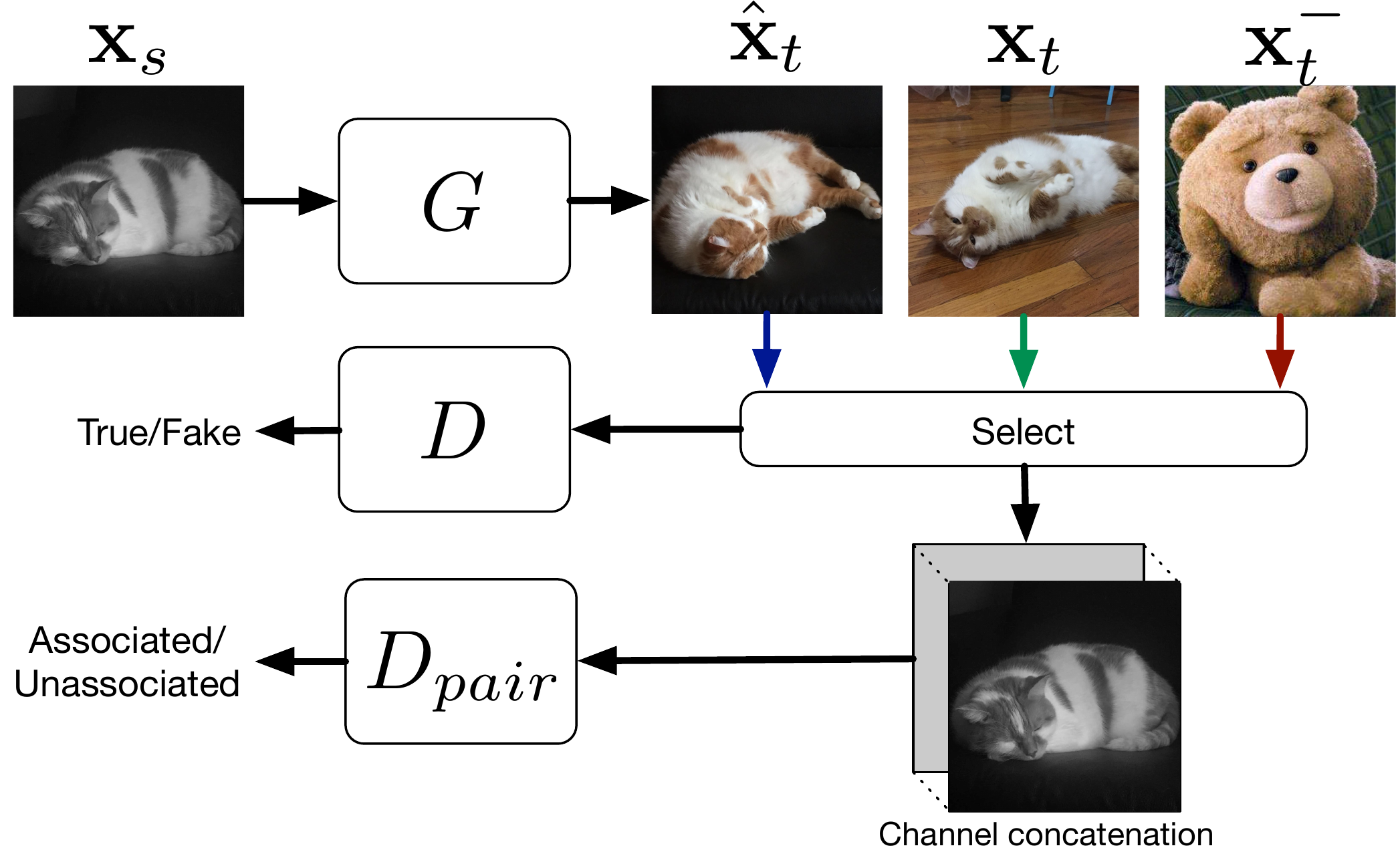}
\caption{Architecture of Pixel-Level Domain Transfer (PLDT)~\cite{yoo2016pldtGAN}. In this example, the source domain is BW color space, while the target domain is RGB space. As explained in \cite{yoo2016pldtGAN}, this model can performs geometric modifications on the input image, while keeping the object in the output image the same as input.}
\label{fig:pldt}
\end{figure}

PLDT~\cite{yoo2016pldtGAN} proposes another method to do supervised \emph{image-to-image translation}, by adding another discriminator $D_{pair}$ that learns to tell whether a pair of images from different domains is associated with each other. The architecture of PLDT is shown in Figure \ref{fig:pldt}. Given an input image $\vec{x}_s$ from source domain, its ground-truth image $\vec{x}_t$ in the target domain,  an irrelevant image $\vec{x}_t^-$ in the target domain, and the generator $G$ transfers  $\vec{x}_s$ into an image $\hat{\vec{x}}_t$ in the target domain,  the loss for $D_{pair}$ can be defined as:
\begin{align}
    \mathcal{L}_{pair} = &\ -t\cdot\log[D_{pair}(\vec{x}_s, \vec{x})] \notag \\
    &\ +(t-1)\cdot\log[1-D_{pair}(\vec{x}_s, \vec{x})], \notag \\
    &\ \textrm{s.t.} \ t= 
    \left\{  
             \begin{array}{lr}  
             0 \quad \textrm{if}\ \vec{x}=\vec{x}_t \\  
             0 \quad \textrm{if}\ \vec{x}=\hat{\vec{x}}_t \\  
             1 \quad \textrm{if}\ \vec{x}=\vec{x}_t^-.
             \end{array}  
\right.  
\end{align}

The generator of PLDT~\cite{yoo2016pldtGAN} is implemented in an encoder-decoder fashion using (transposed) convolutions, while the two discriminators are implemented as fully convolutional networks. As shown in the experimental results, PLDT~\cite{yoo2016pldtGAN} performs domain transfer that modifies the geometric shapes of objects while trying to keep the texture consistent among all associated images.

\begin{figure*}[!htb]
\centering
\includegraphics[width=0.8\linewidth]{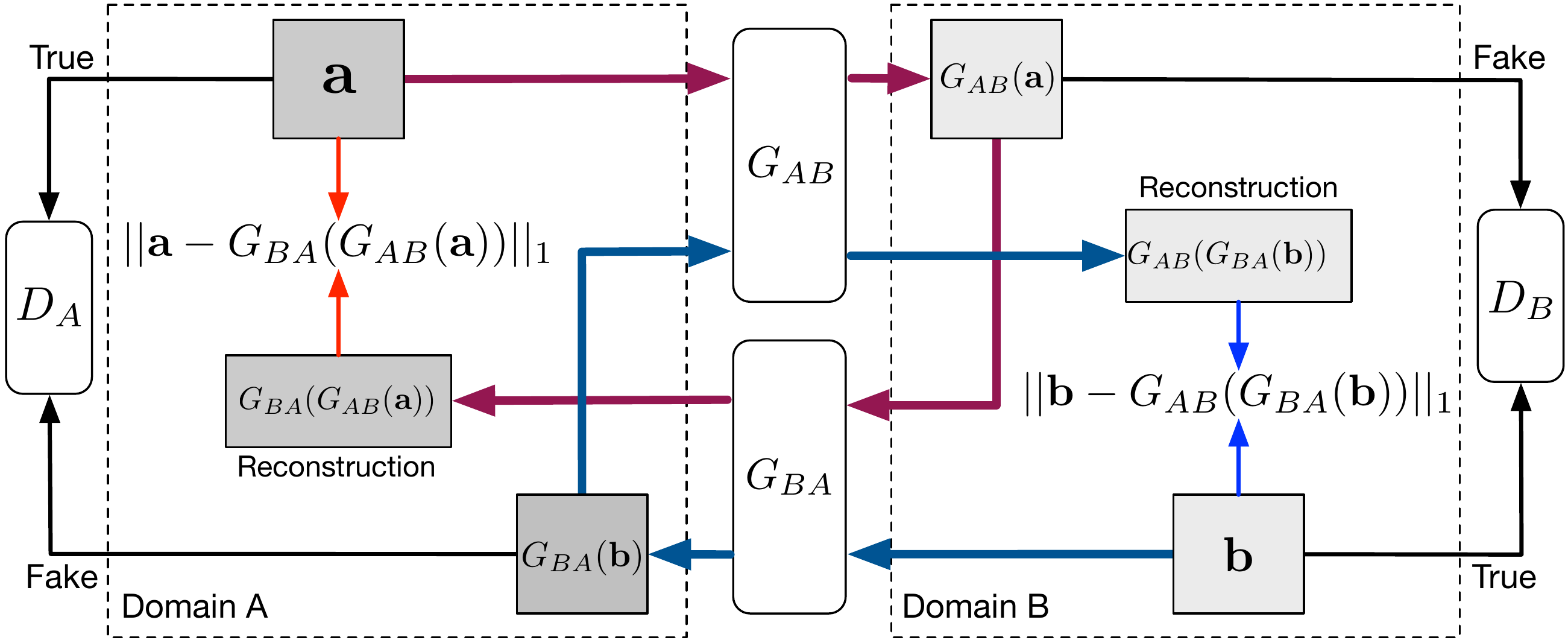}
\caption{Framework of CycleGAN and DualGAN. $A$ and $B$ are two different domains. There is a discriminator for each domain that judges if an image belong to that domain. Two generators are designed to translate an image from one domain to another. There are two cycles of data flow, the red one performs a sequence of domain transfer $A \rightarrow B \rightarrow A$, while the blue one is $B \rightarrow A \rightarrow B$. L1 loss is applied on the input $\vec{a}$ (or $\vec{b}$) and the reconstructed input $G_{BA}(G_{AB}(\vec{a}))$  (or $G_{AB}(G_{BA}(\vec{b}))$) to enforce self-consistency.}
\label{fig:cycleGAN}
\end{figure*}

\subsection{Unsupervised Image-to-Image Translation with Cyclic Loss}
Two concurrent works CycleGAN~\cite{zhu2017cycleGAN} and DualGAN~\cite{yi2017dualGAN} propose to add a self-consistency (reconstruction) loss that tries to preserve the input image after a cycle of transformation. CycleGAN and DualGAN share the same framework, which is shown in Figure~\ref{fig:cycleGAN}. As we can see, the two generators $G_{AB}$ and $G_{BA}$ are doing opposite transformations, which can be seen as a kind of \emph{dual learning}~\cite{he2016dualLearn}. Besides, DiscoGAN~\cite{kim2017discoGAN} is another model that utilizes the same cyclic framework as Figure~\ref{fig:cycleGAN}.

Here we use CycleGAN as an example. In CycleGAN, there are two generators, $G_{AB}$ that transfer an image from domain $A$ to $B$ and $G_{BA}$ that performs the opposite transformation. Also, there are also two discriminators $D_A$ and $D_B$ that predicts whether an image belongs to that domain. For a pair of $G_{AB}$ and $D_B$, the adversarial loss function is defined as:
\begin{align}
\label{eq:cyc-gan-loss}
    \mathcal{L}_{GAN}(G_{AB},D_B) = & \ \mathbb{E}_{\vec{b}\sim p_{B}(\vec{b})}[\log D_B(\vec{b})] \notag \\ 
                                  + & \ \mathbb{E}_{\vec{a}\sim p_{A}(\vec{a})} [1-\log(D_B(G_{AB}(\vec{a}))],
\end{align}
and similarly for the pair $G_{BA}$ and $D_A$, we can define the adversarial loss as $\mathcal{L}_{GAN}(G_{BA},D_A)$. 

Besides the adversarial loss, a cycle-consistency loss is designed to minimize the reconstruction error after we translate an image of one domain to another and then translate it back to the original domain, i.e. $\vec{a} \rightarrow G_{AB}(\vec{a}) \rightarrow G_{BA}(G_{AB}(\vec{a})) \approx \vec{a}$. Since this cycle can be defined from two directions, the cycle-consistency loss is defined as:
\begin{align}
\label{eq:cyc-cyc-loss}
    \mathcal{L}_{cyc}(G_{AB}, G_{BA}) = & \ \mathbb{E}_{\vec{a}\sim p_{A}(\vec{a})}[||\vec{a}-G_{BA}(G_{AB}(\vec{a}))||_1] \notag \\
             + & \ \mathbb{E}_{\vec{b}\sim p_{B}(\vec{b})}[||\vec{b}-G_{AB}(G_{BA}(\vec{b}))||_1].
\end{align}

Then the overall loss function is:
\begin{align}
    \mathcal{L}(G_{AB}, G_{BA},D_A,D_B) = &\ \mathcal{L}_{GAN}(G_{AB},D_B) \notag \\ 
    & + \mathcal{L}_{GAN}(G_{BA},D_A) \notag \\
    & + \lambda\mathcal{L}_{cyc}(G_{AB}, G_{BA}),
\end{align}
where $\lambda$ is a hyper-parameter to balance the losses. Then the objective is to solve for:
\begin{align}
    G_{AB}^*,G_{BA}^* = \textrm{arg} \min_{G_{AB},G_{BA}} \max_{D_B,D_A} \mathcal{L}(G_{AB}, G_{BA},D_A,D_B).
\end{align}

Although CycleGAN~\cite{zhu2017cycleGAN} and DualGAN~\cite{yi2017dualGAN} have the same objective, they use different implementations for generators. CycleGAN uses the generator structure as proposed in \cite{johnson2016perceptual}, while DualGAN follows the U-Net~\cite{ronneberger2015u-net} structure as in \cite{isola2016pix2pix} \cite{ronneberger2015u-net}. Both CycleGAN and DualGAN use the PatchGAN with size $70\times70$ as in \cite{isola2016pix2pix}.

Besides different generator architectures, CycleGAN~\cite{zhu2017cycleGAN} and DualGAN~\cite{yi2017dualGAN} also use different techniques to stabilize the training process. DualGAN follows the training procedure proposed in WGAN~\cite{arjovsky2017wgan}. CycleGAN applies two techniques. First, instead of using the log loss~\cite{mao2017lsgan} for $\mathcal{L}_{GAN}$ in Equation \ref{eq:cyc-gan-loss} with a least square loss that in practice performs more stably and produces higher quality images:
\begin{align}
    \mathcal{L}_{LSGAN}(G_{AB},D_B)= &\ \mathbb{E}_{\vec{b}\sim p_{B}(\vec{b})}[(D_B(\vec{b})-1)^2] \notag \\
                                  + & \ \mathbb{E}_{\vec{a}\sim p_{A}(\vec{a})} [(D_B(G_{AB}(\vec{a}))^2].
\end{align}
The second technique used in CycleGAN~\cite{zhu2017cycleGAN} is that, in order to reduce model oscillation~\cite{goodfellow2014GAN}, CycleGAN follows SimGAN's~\cite{shrivastava2016simGAN} strategy and updates discriminators $D_A$ and $D_B$ using a history of $50$ previously generated images instead of the ones produced by latest generators. 

Experiments of CycleGAN demonstrate the potential of performing high-quality image-to-image translation using unpaired data only, even though supervised method like Pix2Pix~\cite{isola2016pix2pix} still outperforms CycleGAN by a noticeable margin. CycleGAN also conducts experiments that show the importance of using both circles in the circle-consistency loss defined in Equation \ref{eq:cyc-cyc-loss}.  However, failure cases provided in \cite{zhu2017cycleGAN} show that, just as Pix2Pix~\cite{isola2016pix2pix}, CycleGAN does not work in cases that necessitate geometric transformations, such as $apple \leftrightarrow orange$ and $cat \leftrightarrow dog$. More examples are available on CycleGAN's project website\footnote{\url{https://junyanz.github.io/CycleGAN/}}.

\subsection{Unsupervised Image-to-Image Translation with Distance Constraint}
DistanceGAN~\cite{benaim2017distanceGAN} discovers that, the distance $||\vec{x}_i-\vec{x}_j||$ between two images in the source domain $A$ is highly positively correlated to the distance of their counterparts $||G_{AB}(\vec{x}_i)-G_{AB}(\vec{x}_j)||$ in the target domain $B$, which can be seen from Figure 1 of the paper \cite{benaim2017distanceGAN}. According to  \cite{benaim2017distanceGAN}, let $d_k$ be the distance $||\vec{x}_i-\vec{x}_j||$, and $d_k'$ be $||G_{AB}(\vec{x}_i)-G_{AB}(\vec{x}_j)||$, a high correlation indicates that $\sum d_k d_k'$ should also be high. The pair-wise distances $d_k$ in source domain are fixed, and maximizing $\sum d_k d_k'$ causes $d_k$ with large value to dominate the loss, which is undesirable. So the authors propose to minimize $\sum |d_k-d_k'|$ instead.

DistanceGAN~\cite{benaim2017distanceGAN} proposes to use a pair-wise distance loss defined as:
\begin{align}
    \mathcal{L}_{dist}(G_{AB},p_{A}) = &\ \mathbb{E}_{\vec{x}_i,\vec{x}_j\sim p_A} | \frac{1}{\sigma_A}(||\vec{x}_i-\vec{x}_j||_1-\mu_A) \notag \\
     - &\ \frac{1}{\sigma_B}(||G_{AB}(\vec{x}_i)-G_{AB}(\vec{x}_j)||_1-\mu_B)|,
\end{align}
where $\mu_A, \mu_B$ $(\sigma_A,\sigma_B)$ are the pre-computed means (standard deviations) of pair-wise distances in training sets of domain $A,B$ respectively.

In order to support stochastic gradient descent where only one data sample is fed into the model at a time, DistanceGAN proposes another self-distance constraint:
\begin{align}
    \mathcal{L}_{self\textrm{-}dist}(G_{AB},p_{A})&\ = \mathbb{E}_{\vec{x}\sim p_A} | \frac{1}{\sigma_A}(||L(\vec{x})-R(\vec{x})||_1-\mu_A) \notag \\
     - \frac{1}{\sigma_B}&(||G_{AB}(L(\vec{x}))-G_{AB}(R(\vec{x}))||_1-\mu_B)|,
\end{align}
where $L(\vec{x})$ and $R(\vec{x})$ indicate the left and right half of the image $\vec{x}$, and only the left (right) parts of images are taken into account when calculating $\mu_A,\sigma_A$ $(\mu_B, \sigma_B)$. 

Thus the overall loss of DistanceGAN is given by:
\begin{align}
\label{eq:dist-full-loss}
    \mathcal{L} = &\ \alpha_{1A}\mathcal{L}_{GAN}(G_{AB},D_B) + \alpha_{1B}\mathcal{L}_{GAN}(G_{BA},D_A) \notag \\
    &\ +  \alpha_{2A}\mathcal{L}_{dist}(G_{AB},p_A) + \alpha_{2B}\mathcal{L}_{dist}(G_{BA},p_B) \notag \\
    &\ + \alpha_{3A}\mathcal{L}_{self\textrm{-}dist}(G_{AB},p_A) + \alpha_{3B}\mathcal{L}_{self\textrm{-}dist}(G_{BA},p_B) \notag \\
    &\ +  \alpha_{4}\mathcal{L}_{cyc}(G_{AB},G_{BA}),
\end{align}
where $\mathcal{L}_{cyc}(G_{AB},G_{BA})$ is defined in Equation \ref{eq:cyc-cyc-loss} and $\mathcal{L}_{GAN}(\cdot)$ is defined in Equation \ref{eq:cyc-gan-loss}.

DistanceGAN conducts extensive experiments using different losses: $\mathcal{L}_{cyc}$ alone (DiscoGAN \cite{kim2017discoGAN} and CycleGAN \cite{zhu2017cycleGAN}), one-sided distance loss $\mathcal{L}_{dist}$ ($A\rightarrow B \rightarrow A$ or $B \rightarrow A\rightarrow B$), the combination of $\mathcal{L}_{cyc}$ and one-sided $\mathcal{L}_{dist}$, and one-sided self-distance loss $\mathcal{L}_{self\textrm{-}dist}$ alone. The implementation of DistanceGAN~\cite{benaim2017distanceGAN} is the same as baseline (DiscoGAN~\cite{kim2017discoGAN} or CycleGAN~\cite{zhu2017cycleGAN}), dependent on which one is to be compared with.  Results show that the one-sided distance loss or self-distance loss outperforms DiscoGAN and CycleGAN in several tasks, and that the combination of cyclic loss and distance loss achieves the best results in some cases. However, the paper does not explore other possible combinations, such as using both distance losses or even the full loss function defined in Equation \ref{eq:dist-full-loss}.

One interesting thing is, as stated in \cite{benaim2017distanceGAN}, that DistanceGAN computes the distances in raw RGB space and still achieves better performance than baselines, but it may help if the distances are calculated in images' latent feature space where the features can be extracted using pre-trained image classifiers.

In DistanceGAN~\cite{benaim2017distanceGAN}, the authors argue that the high positive correlation between $d_k$ and $d_k'$ implies that $\sum d_k d_k'$ should be high. However, it is unclear how high $d_k'$ should be. For example, if $d_k=1$ and we know that $d_k'$ should be high, we still cannot definitely say that $d_k'=3$ is more desirable than $d_k'=2$. The concept of ``high'' is blurry, so it may be better to use the concept ``higher''. An alternative statement could be, let $\vec{x},\vec{y},\vec{z}$ be images from domain $A$, if $d(\vec{x},\vec{y}) > d(\vec{x},\vec{z})$, then $d(G_{AB}(\vec{x}),G_{AB}(\vec{y})) > d(G_{AB}(\vec{x}),G_{AB}(\vec{z}))$. And thus we can design a loss like $\max (\delta, d(G_{AB}(\vec{x}),G_{AB}(\vec{y})) - d(G_{AB}(\vec{x}),G_{AB}(\vec{z})))$ for all triplets $\vec{x},\vec{y},\vec{z}\in A$ such that $d(\vec{x},\vec{y}) > d(\vec{x},\vec{z})$. 

Regardless of the objective of maximizing $\sum d_k d_k'$, the actual loss used in DistanceGAN, i.e. $\sum |d_k-d_k'|$, is forcing $d_k'$ to be as close to $d_k$ as possible. In other words, how two images of a domain differ from each other should be reflected in the same way when they are translated into another domain, which can be called ``equivariance''. The distance loss and self-distance loss proposed in DistanceGAN~\cite{benaim2017distanceGAN} is essentially capturing this ``equivariance'' property.

\subsection{Unsupervised Image-to-Image Translation with Feature Constancy}
\begin{figure}[htb]
\centering
\includegraphics[width=0.98\linewidth]{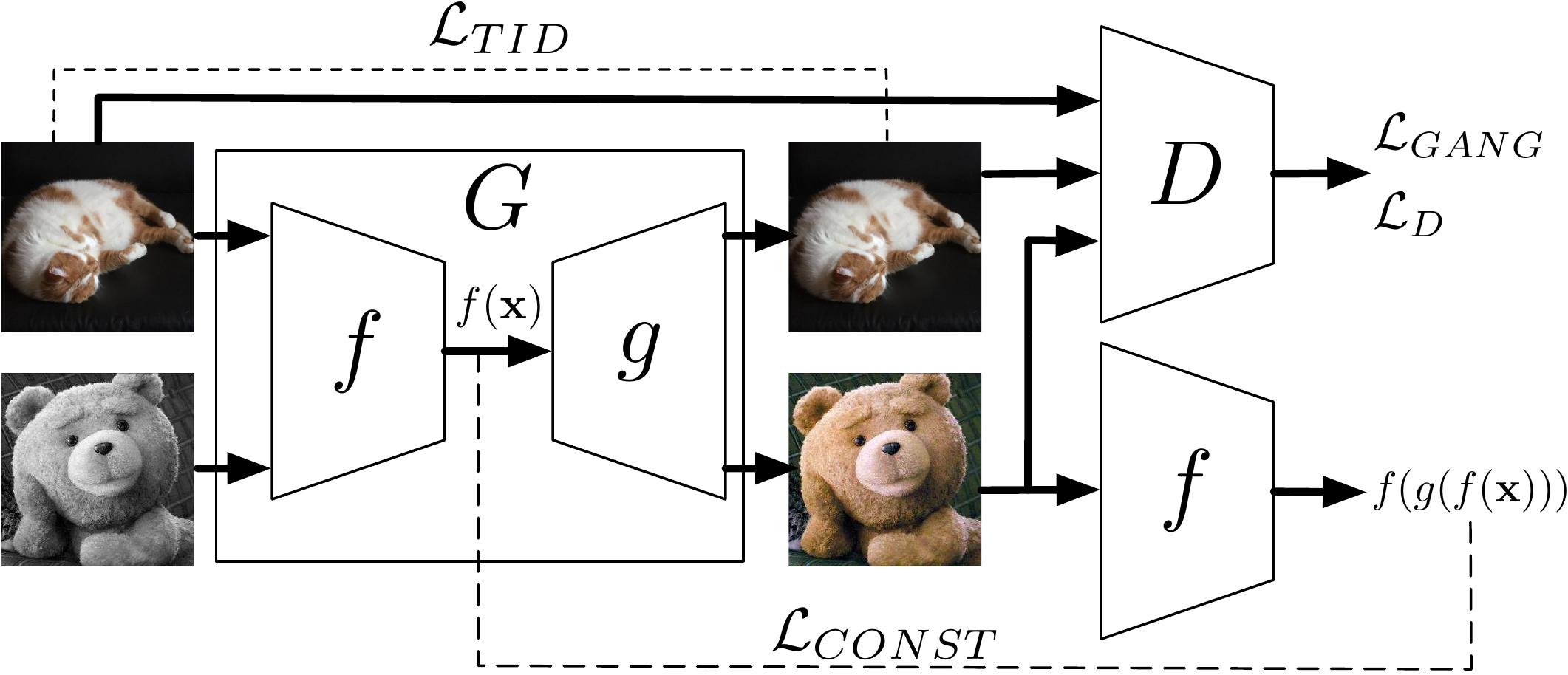}
\caption{Architecture of domain transfer network (DTN). As an example here, the model transfers images from BW color space to RGB color space. The generator is expected to be an identity matrix to images in the target domain, so that in this example, an RGB image remains unchanged when put into the generator, while a BW image is transformed into RGB by the generator. }
\label{fig:dtn}
\end{figure}

As we mention previously, besides minimizing the reconstruction error at raw pixel level, we can also do this at higher feature level, which is explored in DTN~\cite{taigman2016dtn}. The architecture of DTN is shown in Figure~\ref{fig:dtn}, where the generator $G$ is composed of two neural networks, a convolutional network $f$ and an transposed convolutional network $g$ such that $G=g\circ f$. 

Here $f$ acts as a feature extractor, and DTN~\cite{taigman2016dtn} tries to preserve high level features of an input image from source domain after it is transferred into target domain. Let $X_s$ denote source domain and $X_t$ be target domain. Given an input image $\vec{x}\in X_s$, the output of generator is $G(\vec{x})=g(f(\vec{x}))$, then the feature reconstruction error can be defined with a distance measure $d$ (DTN uses mean squared error (MSE)):
\begin{align}
    \mathcal{L}_{CONST} = \sum_{\vec{x}\in X_s}d(f(\vec{x}, f(g(f(\vec{x}))))).
\end{align}

Besides, DTN also expects the generator $G$ to act as an identity matrix to images from the target domain. Given a distance measure $d_2$ (DTN uses mean squared error (MSE)), an identity mapping loss for target domain is defined as:
\begin{align}
    \mathcal{L}_{TID} = \sum_{\vec{x}\in X_t}d_2(\vec{x}, g(f(\vec{x}))).
\end{align}

In addition, DTN use a multi-class discriminator instead of a binary one in regular GAN. Here $D$ is a ternary classification function that maps an image to one of the three classes \{1,2,3\}, where class 1 means that the image is from source domain but transformed by $G$, class 2 means that the input is an image from target domain but transformed by $G$, and class 3 means that the input image is from target domain without any transformation. Let $D_i(\vec{x})$ denotes the probability of $\vec{x}$ belongs to class $i$, the discriminator loss $\mathcal{L}_D$ is defined as:
\begin{align}
    \mathcal{L}_D = &\ -\mathbb{E}_{\vec{x}\in X_s}\log D_1(g(f(\vec{x}))) \notag \\ 
     &\ -\mathbb{E}_{\vec{x}\in X_t}\log D_2(g(f(\vec{x}))) -\mathbb{E}_{\vec{x}\in X_t}\log D_3(\vec{x}).
\end{align}
Similarly, the generator's adversarial loss $\mathcal{L}_{GANG}$ is defined as:
\begin{align}
    \mathcal{L}_{GANG} = &\ -\mathbb{E}_{\vec{x}\in X_s}\log D_3(g(f(\vec{x}))) \notag \\ 
     &\ -\mathbb{E}_{\vec{x}\in X_t}\log D_3(g(f(\vec{x}))).
\end{align}
In order to slightly smoothen the generated images, DTN adds an anisotropic total variation loss~\cite{mahendran2015under-invert} $\mathcal{L}_{TV}$ defined on generated images $\vec{z}=[z_{ij}]=G(\vec{x})$:
\begin{align}
    \mathcal{L}_{TV}(\vec{z})=\sum_{i,j}((\vec{z}_{i,j+1}-\vec{z}_{ij}^2 + (\vec{z}_{i+1,j}-\vec{z}_{ij})^2)^{\frac{1}{2}}.
\end{align}
Then the overall generator loss $\mathcal{L}_G$ is defined as:
\begin{align}
    \mathcal{L}_G = \mathcal{L}_{GANG} + \alpha\mathcal{L}_{CONST} + \beta\mathcal{L}_{TID} + \gamma\mathcal{L}_{TV}.
\end{align}

Experiments show that DTN~\cite{taigman2016dtn} produces impressive images on face-to-emoji task, which is competitive with some existing emoji generating programs.

\subsection{Unsupervised Image-to-Image Translation with Auxiliary Classifier}
Bousmalis et. al~\cite{bousmalis2016upldaGAN} propose to use a task-specific auxiliary classifier to help unsupervised \emph{image-to-image translation}. We denote their model as DAAC (short for ``Domain Adaption with Auxiliary Classifier'') here for convenience. DAAC~\cite{bousmalis2016upldaGAN} contains a task-specific classifier $C(\vec{x})\rightarrow \vec{y}$ that assigns a label vector $\vec{y}$ to an image in either source domain or target domain. This classifier $C$, for example, can be an image classification model. The architecture of DAAC is similar to Figure \ref{fig:acgan}.

Given a training set $\mathbf{X}_s$ in which each image $\mathbf{x}\in \mathbf{X}_s$ has a class label $\mathbf{y}_\mathbf{x}$, the objective of $C$ is to minimize the cross-entropy loss $\mathcal{L}_C$:
\begin{align}
    \mathcal{L}_{c}(G,C) = \mathbb{E}_{\mathbf{x} \in \mathbf{X}_s}[-\mathbf{y}_\mathbf{x}^{\intercal} \log[C(G(\mathbf{x}))] -\mathbf{y}_\mathbf{x}^{\intercal}\log[C(\mathbf{x})]].
\end{align}

Besides, DAAC~\cite{bousmalis2016upldaGAN} also proposes a content-similarity loss in cases with prior knowledge on what information should be preserved after the domain adaption process. For example, we may expect the hues of the source image and the adapted image to be the same. In \cite{bousmalis2016upldaGAN}, the authors consider the case where they render objects with black background and expect the adapted images to have the same objects but different backgrouds. In this case, each image $\mathbf{x}$ is associated with a binary mask $\mathbf{m}_\mathbf{x} \in \mathbb{R}^k $ ($k$ is the number of pixels in $\mathbf{x}$) to separate foreground and background, and the content-similarity loss $\mathcal{L}_s$, which is a variation of the pairwise mean squared error (PMSE)~\cite{eigen2014pmse}, can be defined as:
\begin{align}
    \mathcal{L}_s(G) = &\ \mathbb{E}_{\mathbf{x}\in \mathbf{X}_s} [\frac{1}{k}||(\mathbf{x}-G(\mathbf{x}))\circ \mathbf{m}_{\mathbf{x}}||^2_2 \notag \\
    &\ -\frac{1}{k^2} (( \mathbf{x}-G(\mathbf{x}))^{\intercal} \mathbf{m}_{\mathbf{x}})^2 ],
\end{align}
where $||\cdot||^2_2$ is the squared L2 norm, and $\circ$ is the Hardamard product. Note that here the image $x$ is flattened into a vector form.

The total objective is thus given by:
\begin{align}
     \textrm{arg} \min_{G,C} \max_{D} \alpha \mathcal{L}_{GAN}(G,D) + \beta \mathcal{L}_c(G,C) + \gamma \mathcal{L}_s(G),
\end{align}
where $\mathcal{L}_{GAN}(G,D)$ is the regular GAN objective defined in Equation~\ref{eq:gan_obj}, and $\alpha, \beta,\gamma$ are hyper-parameters to balance different terms of the objective.

Although $\mathcal{L}_c(G,C)$ and $\mathcal{L}_s(G)$ help in generating better adapted images, the amount of labeled images is limited in real world, and fine-grained pixel-level masks are even harder to obtain. Nonetheless, using auxiliary classifier can also help to avoid \emph{mode collapse}, as in AC-GAN~\cite{odena2016acGAN}.

\begin{figure}[!htb]
\centering
\includegraphics[width=0.7\linewidth]{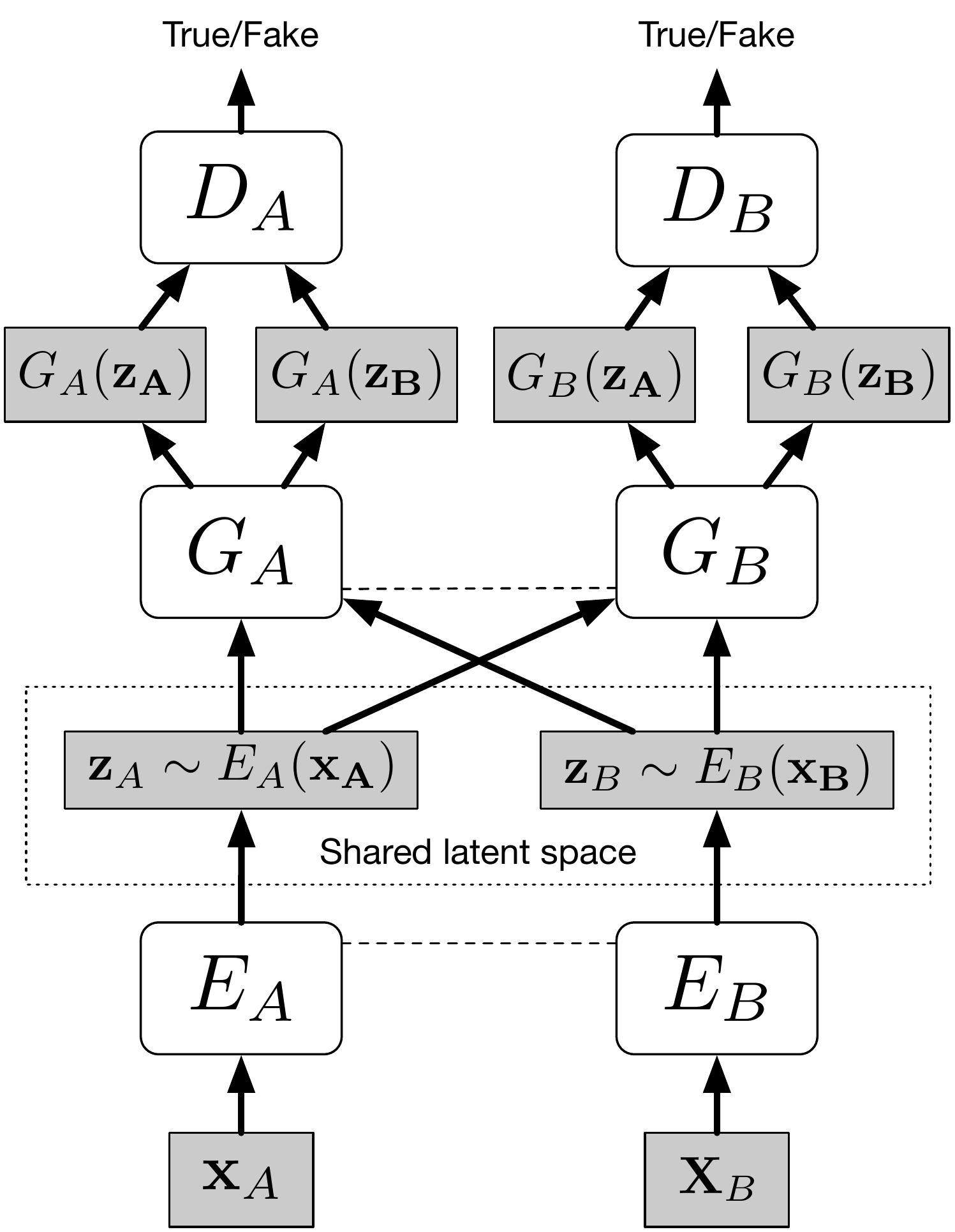}
\caption{Architecture of UNIT. The two encoders share weights at the last few layers, while the two generators share weights at the first few layers, as indicated by the dashed lines. }
\label{fig:unit}
\end{figure}

\subsection{Unsupervised Image-to-Image Translation with VAE and Weight Sharing}
UNIT~\cite{liu2017unit} proposes to add VAE to CoGAN~\cite{liu2016coGAN} for unsupervised image-to-image translation, as illustrated in Figure \ref{fig:unit}. In addition, UNIT assumes that both encoders share the same latent space, which means, let $\vec{x}_A, \vec{x}_B$ be the same image in different domains, and then the shared latent space implies that $E_A(\vec{x}_A) = E_B(\vec{B}_B)$. Based on the shared-latent space assumption, UNIT enforces weight sharing between the last few layers of the encoders and between the first few layers of the generators. The objective function for UNIT is a  combination of the objectives of GAN and VAE, with the difference of using two sets of GANs/VAEs and adding hyper-parameters $\lambda$s to balance different loss terms. Also, UNIT states that the shared latent space assumption implies the cycle-consistency~\cite{zhu2017cycleGAN}~\cite{kim2017discoGAN}~\cite{yi2017dualGAN}, so it adds another constraint to its objective which is VAE-like:
\begin{align}
    \mathcal{L}_{cc1} = & \lambda_3 D_{KL}(q_A(\vec{z}_A|\vec{x}_A)||p_{\eta}(\vec{z})) \\ \notag
    & + \lambda_3 D_{KL}(q_B(\vec{z}_B|F_{AB}(\vec{x}_A))||p_{\eta}(\vec{z})) \\ \notag
    & - \lambda_4 \mathbb{E}_{\vec{z}_B\sim q_B(\vec{z}_B|F_{AB}(\vec{x}_A))}[\log p_{G_A}(\vec{x}_A|\vec{z}_B)],
\end{align}
where $q_A(\vec{z}_A|\vec{x}_A) = \mathcal{N}(\vec{z}_A|E_A(\vec{x}_A), \mathbf{I})$, $q_B(\vec{z}_A|\vec{x}_B) =  \mathcal{N}(\vec{z}_B|E_A(\vec{x}_B), \mathbf{I})$, $p_{\eta}(\vec{z}) = \mathcal{N}(\vec{z}|0, \mathbf{I})$, and $F_{AB}(\vec{x})=G_B(E_A(\vec{x}))$. And $\mathcal{L}_{cc2}$ can be defined similarly by reversing subscripts of $A$ and $B$.

Although UNIT performs better than models like DTN~\cite{taigman2016dtn} and CoGAN~\cite{liu2016coGAN} on MNIST~\cite{lecun1998mnist}, Street View House Number (SVHN)~\cite{netzer2011SVHN} datasets in terms of cross-domain classification accuracy, it does not compare with other unsupervised methods such as CycleGAN~\cite{zhu2017cycleGAN} and  DiscoGAN~\cite{kim2017discoGAN}, nor does it use other widely adopted evaluation metrics like \emph{Inception Score}.

\subsection{Unsupervised Multi-domain Image-to-Image Translation}
Previous models can only transform images between two domains, but if we want to transform an image among several domains, we need to train a separate generator for each pair of domains, which is costly. To deal with this problem, StarGAN~\cite{choi2017stargan} proposes to use one generator that can generate images of all domains. Instead of taking only an image as conditional input, StarGAN also takes the label of target domain as input, and the generator is designed to transform the input image to the target domain indicated by the input label. Similar to DAAC~\cite{bousmalis2016upldaGAN} and AC-GAN~\cite{odena2016acGAN}, StarGAN uses an auxiliary domain classifier which classifies an image into its belonged domain. In addition, a cycle-consistency loss~\cite{zhu2017cycleGAN} is used to preserve the content similarity between input and output images. In order to allow StarGAN to train on multiple datasets that may have different sets of labels, StarGAN uses an additional one-hot vector to indicate the datset and concatenates all label vectors into one vector, setting the unspecified labels as zero for each dataset.

\subsection{Summary on General Image-to-Image Translation}
So far we have discussed some general \emph{image-to-image translation} methods, the different losses they use are summarized in Table~\ref{tab:img2img-sum}. 

The simplest loss is the \emph{pixel-wise L1 reconstruction loss} operates in the target domain, which requires paired training samples. Both \emph{one-sided} and \emph{bi-directional} reconstruction loss can treated as the unsupervised version of \emph{pixel-wise L1 reconstruction loss}, since they enforce cycle-consistency and do not require paired training samples. The additional \emph{VAE loss} is based on the assumption of shared latent space of both source and target domain, and it also implies the bi-directional cycle-consistency loss. The \emph{equivariance loss}, however, does not try to reconstruct images, but to preserve the difference between images across source and target domain. 

Different from previously mentioned losses that work on the generators directly, \emph{pair-wise discriminator loss}, \emph{ternary discriminator loss} and \emph{auxiliary classifier loss} work on the discriminator side and make the discriminator better at distinguishing real and fake samples, and then the generator can also learn better with the enhanced discriminator.

\begin{table}[!htb]
\centering
\caption{Summary of losses used in image \emph{image-to-image translation}, ``sup'' is short for ``supervision''. ``$\surd$'' in both ``source domain'' and ``target domain'' columns indicates that the loss requires images from both domains (but not necessarily paired).}
\label{tab:img2img-sum}
\begin{tabular}{|l|c|c|l|l|}
\hline
\textbf{Loss} & \multicolumn{1}{l|}{\textbf{\begin{tabular}[c]{@{}l@{}}Source \\ domain\end{tabular}}} & \multicolumn{1}{l|}{\textbf{\begin{tabular}[c]{@{}l@{}}Target \\ domain\end{tabular}}} & \textbf{Sup} & \textbf{Models} \\ \hline
Pixel-level L1 &  & $\surd$ & \multicolumn{1}{c|}{$\surd$} & Pix2Pix\cite{isola2016pix2pix} \\ \hline
\begin{tabular}[c]{@{}l@{}}Pair-wise \\ discriminator\end{tabular} & \multicolumn{1}{l|}{} & $\surd$ & \multicolumn{1}{c|}{$\surd$} & PLDT\cite{yoo2016pldtGAN} \\ \hline
\begin{tabular}[c]{@{}l@{}}Bi-directional\\ reconstruction\end{tabular} & $\surd$ & $\surd$ &  & \begin{tabular}[c]{@{}l@{}}CycleGAN\cite{zhu2017cycleGAN}, \\ DualGAN\cite{yi2017dualGAN},\\ DiscoGAN\cite{kim2017discoGAN},\\ StarGAN\cite{choi2017stargan}\end{tabular} \\ \hline
\begin{tabular}[c]{@{}l@{}}One-sided \\ reconstruction\end{tabular} & $\surd$ &  &  & DTN\cite{taigman2016dtn} \\ \hline
\begin{tabular}[c]{@{}l@{}}Ternary \\ discriminator\end{tabular} & \multicolumn{1}{l|}{} & $\surd$ &  & DTN\cite{taigman2016dtn} \\ \hline
Equivariance & $\surd$ & $\surd$ &  & DistanceGAN\cite{benaim2017distanceGAN} \\ \hline
\begin{tabular}[c]{@{}l@{}}Auxiliary \\ classifier\end{tabular} & $\surd$ & $\surd$ &  & \begin{tabular}[c]{@{}l@{}}DAAC\cite{bousmalis2016upldaGAN},\\ StarGAN\cite{choi2017stargan}\end{tabular} \\ \hline
VAE & $\surd$ & $\surd$ &  & UNIT\cite{liu2017unit} \\ \hline
\end{tabular}
\end{table}

Among all mentioned models, Pix2Pix~\cite{isola2016pix2pix} produces sharpest images, even though the L1 loss is just a simple add-on component to the original GAN model. It may be interesting to combine L1 loss with the pair-wise discriminator in PLDT~\cite{yoo2016pldtGAN} which may improve the model's performance on image-to-image translations that involve geometric changes on images. Also, Pix2Pix may benefit from preserving similarity information between images in source and target domains, as done in some unsupervised methods like CycleGAN~\cite{zhu2017cycleGAN} and DistanceGAN\cite{benaim2017distanceGAN}. As for unsupervised methods, although their results are not as sharp as supervised methods like Pix2Pix~\cite{isola2016pix2pix}, they are a promising research direction, since they do not require paired data and collecting labeled data is very costly in real world.

\subsection{Task-Specific Image-to-Image Translation}
In this section, we are going to introduce some models that work on more specific \emph{image-to-image translation} applications.

\subsubsection{Face Editing}
\emph{Face editing} is closely related to \emph{image-to-image translation}, since it also takes images as input and produces images.  However, \emph{Face editing} focuses more on manipulating the attributes of humans' faces while \emph{image-to-image translation} is a more general scope. 

IcGAN~\cite{perarnau2016icGAN} proposes to learn two separate encoders, $E_\vec{z}$ that maps an image to its latent vector $\vec{z}$ and $E_\vec{y}$ that learns the attribute information vector $\vec{y}$. Attribute manipulation is performed by tuning the attribute vector $\vec{y}$, concatenating it with $\vec{z}$ and then putting the combined vector as input to the generator.

Instead of generating images directly, Shen et. al~\cite{ShenL2016res-face} propose to learn \emph{residual} images of attributes. To add an attribute to an image $\vec{x}$,  a residual image  $G(\vec{x})$ is obtained by putting it through a generator $G$, and then the manipulated image is obtained by $\vec{x} + G(\vec{x})$. The framework of this model follows the \emph{dual learning} approach we discuss earlier, and the discriminator is similar to the ternary classifier in DTN~\cite{taigman2016dtn}, which distinguishes whether the image is from ground-truth or is manipulated by either of the two generators ($G_{AB}, G_{BA}$).

Recently, Brock et. al propose Introspective Adversarial Network (IAN)~\cite{brock2016IAN} for neural photo editing, which, similar to VAE-GAN~\cite{larsen2015vae-gan}, is also a combination of GAN~\cite{goodfellow2014GAN} and VAE~\cite{kingma2013vae}. However, unlike VAE-GAN that uses different networks for discriminator and encoder, IAN combines the discriminator and encoder into a single network. The intuition behind it is that features learned by a well trained discriminator tend to be more expressive than those learned by maximum likelihood, which is the reason why they are more suitable for inference. Specifically, the encoder is implemented as a fully-connected layer on top of the last convolution layer of the discriminator.

Besides manipulating face attributes, there are also other forms of ``face editing'' such as generating the frontal view of a person's face based on pictures of the face's side view. Huang et. al propose Two-Pathway Generative Adversarial Network (TP-GAN)~\cite{Huang2017TP-GAN} that performs such task. TP-GAN consists of two pathways, a global structure pathway and a local texture pathway, where each pathway contains a pair of encoder and decoder. The global pathway constructs a blurry frontal view that captures the global structure of the face, while the local pathway with four sub-networks attends to local texture details around four facial landmarks, which are \emph{left eye center}, \emph{right eye center}, \emph{nose tip} and \emph{mouth center}. Outputs of four sub-networks in the local pathway are concatenated with the output of global pathway, and then the combined tensor is fed into successive convolution layers to produce the final synthetic image. In addition to L1 pixel loss, TP-GAN proposes to use a symmetry loss to constrain the symmetry of human faces, and an identity preserving loss to preserve the identity of input face. The identity loss is implemented as a perceptual loss~\cite{johnson2016perceptual}.

\subsubsection{Image Super-Resolution}
Another application of GAN that is related to \emph{image-to-image translation} is image super-resolution, which is the task of taking a low resolution image as input and outputs a high resolution one with sharp details. SRGAN~\cite{Ledig2016srgan} proposes to use a residual block~\cite{he2016resnet} based generator and a fully convolutional discriminator~\cite{radford2015dcgan} to do single image super-resolution. Besides adversarial loss, SRGAN also combines pixel-wise MSE loss, perceptual loss~\cite{johnson2016perceptual} and regularization loss~\cite{johnson2016perceptual}. SRGAN outperforms several baselines on some metrics by a small margin, but the difference in synthetic images is not easy to tell without zooming in.

\subsubsection{Video Prediction}
A special kind of related application is \emph{video prediction}, which aims to predict the next frame of a video given the current frame (or a history of frames). 

VGAN~\cite{vondrick2016vgan} proposes to use a hierarchical model for video prediction. The generator has two data streams, a foreground stream $f$ consisting of 3D transposed convolutions, and a background stream $b$ consisting of 2D transposed convolutions. The foreground stream is also responsible for generating a mask $m$ that is used to merged the results of two streams: $m\odot f + (1-m)\odot b$. The discriminator is a set of spatial-temporal convolution layers.

Mathieu et. al propose Adv-GDL~\cite{mathieu2015adv-gdl} which generates future frames in an iterative way, from low to high resolution. Let $s_1,s_2,...,s_k$ be a set of sizes and $u_k$ be the upsampling operator from size $s_{k-1}$ to $s_k$. Let $\vec{X}_k, \vec{Y}_k$ be ground-truth current and future frames of size $s_k$ and $G_k$ be the generator that produces images of size $s_k$,then the predicted future frame $\hat{\vec{Y}}_k$ is calculated by $\hat{\vec{Y}}_k = u_k(\hat{\vec{Y}}_{k-1}) + G_k(\vec{X}_k,u_k(\hat{\vec{Y}}_{k-1}))$. The discriminator is a series of multi-scale convolutional networks with scalar output. 

Although both methods produce reasonable output videos when the time interval is short, video quality becomes worse as the time increases. Objects in synthetic videos lose their original shapes and get morphed to indistinguishable objects in some cases, which may be due to the models' inability to learn legal movements of objects.

\section{Evaluation Metrics on Synthetic Images}
\label{sec:metrics}
It is very hard to quantify the quality of synthetic images, and metrics like RMSE are not suitable since there is no absolute one-to-one correspondence between synthetic and real images. A commonly used subjective metric is to use the Amazon Mechanical Turk (AMT)\footnote{\url{https://www.mturk.com/}} that hires humans to score synthetic and real images according to how realistic they think the images are. However, people often have different opinions of what is good or bad, so we also need objective metrics to evaluate the quality of images. 

\emph{Inception score} (IS)~\cite{salimans2016improvedGAN} evaluates an image based on the entropy in class probability distribution when it is put into a pre-trained image classifier. One intuition behind \emph{Inception score} is that the better an image $\vec{x}$ is, the lower the entropy of conditional distribution $p(y|\vec{x})$  should be, which means the classifier have high confidence of what the image is about. Also, to encourage the model to generate various classes of images, the marginal distribution $p(y)=\int p(y|\vec{x}=G(\vec{z}))d\vec{z}$ should have high entropy. Combining these two intuition, the \emph{Inception score} is calculated by $\textrm{exp}(\mathbb{E}_{\vec{x}\sim G(\vec{z})}D_{KL}(p(y|\vec{x})||p(y))$. As discussed in \cite{lucic2017gan-equal}, \emph{Inception score} is neither sensitive to prior distribution of labels, nor a proper distance measure. Also, \emph{Inception score} suffers from \emph{intra-class mode collapse}, since a model only needs to generate one perfect sample for each class to get a perfect \emph{Inception score}.

Similar to \emph{Inception score}, \emph{FCN-score}~\cite{isola2016pix2pix} adopts the idea that if the synthetic images are realistic, classifiers trained on real images will be able to classify the synthetic images correctly. However,  an image classifier does not require the input image to be very sharp as to give a correct classification, which means that metrics based on image classifier may not be able to tell between two images with only small difference in details. Worse still, research in adversarial examples~\cite{goodfellow2014explaining} shows that the decision of a classifier does not necessarily depend on visual content of images but can be highly influenced by noise invisible to humans, which raises more questions on this metric. 

\emph{Fr\textrm{$\acute{e}$}chet Inception Distance} (FID)~\cite{heusel2017fid} provides a different approach. First, generated images are embedded into a latent feature space of a chosen layer of the Inception Net. Second, embeddings of generated and real images are treated as samples from two continuous multivariate Gaussians so that their means and covariances can be calculated. Then the quality of generated images can be determined by the Fr\textrm{$\acute{e}$}chet Distance between the two Gaussians:
\begin{align}
    \textrm{FID}(x,g) = ||\mu_x - \mu_g||^2_2 + \textrm{Tr}(\Sigma_x + \Sigma_g - 2(\Sigma_x\Sigma_g)^{\frac{1}{2}}),
\end{align}
where $(\mu_x,\mu_g)$ and $(\Sigma_x, \Sigma_g)$ are the means and covariances of the samples from the true data distribution and generator's learned distribution respectively. The authors of \cite{heusel2017fid} show that FID is consistent with human judgment and that there is a strong negative correlation between FID and the quality of generated images. Furthermore, FID is less sensitive to noise than IS and can detect intra-class mode collapse.

Besides \emph{Inception score} (IS), \emph{FCN-score} and \emph{Fr\textrm{$\acute{e}$}chet Inception Distance} (FID), there are also other metrics like \emph{Gaussian Parzen Window}~\cite{goodfellow2014GAN}, \emph{Generative Adversarial Metric } (GAM)~\cite{ImKJM2016gran} and \emph{MODE Score}~\cite{che2016mrGAN}. Among all these metrics, \emph{Inception score} is the most widely adopted one for quantitatively evaluating synthetic images, and there is a recent study~\cite{fedus2017gan-path} that uses IS to compare several GAN models. Although FID is relatively new, it has been shown to be better than IS~\cite{heusel2017fid}\cite{lucic2017gan-equal}.

\section{Discriminators as Learned Loss Functions}
\label{sec:learn_loss}
Generative adversarial network (GAN) is powerful and effective in that the discriminator acts as a learned loss function instead of a fixed one designed carefully for each specific task. This is particularly important for image synthesis tasks whose loss functions are hard to be explicitly defined in math. For example, in style transfer task, it is hard to write down a math equation that evaluates how well an image matches a certain painting style. For image synthesis tasks, each input may have many legal outputs, but samples in training set cannot cover all situations. In this case, it is inappropriate to only minimize the distance between synthetic and ground-truth images, since we want the generator to learn the data distribution instead of remembering training samples. Although we can design feature-based losses that try to preserve feature consistency instead of at raw pixel level, as done in the \emph{perceptual loss}~\cite{johnson2016perceptual} for image style, such losses are constrained by pre-trained image classification models they use, and it remains a question of which layers to pick for calculating feature loss  when we switch to another pre-trained model. A discriminator, on the other hand, does not require explicit definition of the loss, since it learns how to evaluate a data sample as it trains against the generator. Thus the discriminator is able to learn a better loss function given enough training data. 

The fact that the discriminator acts as a learned loss function has significant meaning for general artificial intelligence. Traditional pattern recognition and machine learning require us to define what features to be used (e.g. SIFT~\cite{lowe1999sift} and HOG~\cite{dalal2005hog} descriptors), and we design specific loss functions and decide what optimization methods to be applied. Deep learning free us from carefully designing features, by learning low-level and high-level feature representations by itself during training (e.g. CNN kernels), but we still need to work hard at designing loss functions that work well. GAN takes us one step forward on our path towards artificial intelligence, in that it learns how to evaluate data samples instead of being told how to do so, although we still need to design the adversarial loss and combine it with other auxiliary losses. In other words, previously we design how to calculate how close an output is to the corresponding ground-truth ($\mathcal{L}(\vec{x},\hat{\vec{x}})$), but the discriminator learns how to calculate how well an output matches the true data distribution ($\mathcal{L}(\vec{x})$). Such property allows models to be more flexible and more likely to generalize well. Furthermore, with learn2learn~\cite{andrychowicz2016learn2learn} which allows neural networks to learn to optimize themselves, there is a possibility that we may no longer need to choose what optimizers (such as RMSprop~\cite{tieleman2012rmsprop}, Adam~\cite{kingma2014adam} etc.) to use and let models handle everything themselves.

\section{Discussion and Conclusion}
\label{sec:conclusion}
In this paper, we review some basics of Generative Adversarial Nets (GAN)~\cite{goodfellow2014GAN}, and classify image synthesis methods into three main approaches, i.e. \emph{direct method}, \emph{hierarchical method} and \emph{iterative method}, and mention some other generation methods such as iterative sampling~\cite{nguyen2016ppgn}. We also discuss in details two main forms of image synthesis, i.e. \emph{text-to-image synthesis} and \emph{image-to-image translation}.

For \emph{text-to-image synthesis}, current methods work well on datasets where each image contains single object such as CUB~\cite{wah2011cub_bird} and Oxford-102~\cite{Nilsback08oxfordFlower}, but the performance on complex datasets such as MSCOCO~\cite{lin2014mscoco} is much worse. Although some models can produce realistic images of rooms in LSUN~\cite{yu15lsun}, it should be noted that rooms do not contain living things, and a living thing is certainly much more complicated than static objects. This limitation probably stems from the models' inability to learn different concepts of objects. We also propose that one possible way to improve GAN's performance in this task is to train different models that generate single object well and train another model that learns to combine different objects according to text descriptions, and that CapsNet~\cite{sabour2017capsnet} may be useful in such tasks.

For \emph{image-to-image translation}, we review some general methods from supervised to unsupervised settings, such as pixel-wise loss~\cite{isola2016pix2pix}, cyclic loss~\cite{zhu2017cycleGAN} and self-distance loss~\cite{benaim2017distanceGAN}. Besides, we also introduce some task-specific image-to-image translation models for face editing, video prediction and image super-resolution. Image-to-image translation is certainly an interesting application of GAN, which has great potential to be incorporated into other software products, especially mobile apps. Although research in unsupervised methods seems more popular, supervised methods may be more practical since they still produce better synthetic images than unsupervised methods.

Finally, we review some evaluation metrics for synthetic images, and  discuss GAN's role on our path towards artificial intelligence. The power of GAN largely lies in its discriminator's acting as a learned loss function, which makes the model perform better on tasks whose output is hard to evaluate by designing an explicit math equation.

\ifCLASSOPTIONcompsoc
  \section*{Acknowledgments}
\else
  \section*{Acknowledgment}
\fi

He Huang would like to thank Chenwei Zhang and Bokai Cao for the valuable discussions and feedback.

\ifCLASSOPTIONcaptionsoff
  \newpage
\fi



%

\bibliographystyle{IEEEtran}
\bibliography{reference}


%


\begin{IEEEbiographynophoto}{He Huang}
is a second-year PhD student from the department of Computer Science, University of Illinois at Chicago, USA. He received his bachelor's degree in Software Engineering from Sun Yat-sen University, China. His research interests are GAN, computer vision and data mining, especially cross-model tasks (such as image captioning, cross-model retrieval and text-to-image). He is a Bayesian person, and loves probabilistic graphical models for their elegance. 
\end{IEEEbiographynophoto}


\begin{IEEEbiographynophoto}{Phillip S. Yu} is a Disthinguished Professor in the Department of Computer Science at UIC and also holds the Wexler Chair in Information and Technology. Before joining UIC, he was with IBM Thomas J. Watson Research Center, where he was manager of the Software Tools and Techniques department. His main research interests include big data, data mining (especially on graph/network mining), social network, privacy preserving data publishing, data stream, database systems, and Internet applications and technologies.
\end{IEEEbiographynophoto}

\begin{IEEEbiographynophoto}{Changhu Wang} is currently a technical director of Toutiao AI Lab, Beijing, China. He is leading the visual computing team in Toutiao AI Lab, working on computer vision, multimedia analysis, and machine learning. Before joining Toutiao AI Lab, he worked as a lead researcher in Microsoft Research from 2009 to 2017. He was a research engineer at the department of Electrical and Computer Engineering in National University of Singapore in 2008. 
\end{IEEEbiographynophoto}




\end{document}